%% file: main.tex
\pdfoutput=1
\documentclass[11pt,a4paper,onecolumn]{googledeepmind}
\usepackage{multirow}
\usepackage{makecell}
\usepackage{fvextra}
\usepackage{placeins}
\usepackage{flafter}
\usepackage{float}
\usepackage{fontawesome5}
\usepackage[hyphens]{xurl}
\usepackage[numbers,sort&compress]{natbib}

\usepackage[capitalize,noabbrev,nameinlink]{cleveref}
\definecolor{ewmblue}{HTML}{244D6B}
\colorlet{blue}{ewmblue}
\renewcommand{\titlefont}{\color{ewmblue}\normalfont\bfseries\fontsize{21}{23}\selectfont}
\renewcommand{\headingfont}{\color{ewmblue}\normalfont\bfseries\fontsize{13}{14}\selectfont}
\hypersetup{colorlinks=true,allcolors=ewmblue,
  pdftitle={Esports World Model Dataset},
  pdfauthor={Weiyu Ma}}
\renewcommand{\absfont}{\normalfont\fontsize{10}{12}\selectfont}
\newcolumntype{Y}{>{\raggedright\arraybackslash}X}
\newcolumntype{P}[1]{>{\raggedright\arraybackslash}p{#1}}
\fancypagestyle{firststyle}{
  \fancyhf{}
  \fancyhead[L]{\headerfont Preprint}
  \fancyhead[R]{\headerfont October 2026}
    \fancyfoot[L]{\footerfont $^{\dagger}$Project leader: Weiyu Ma, \href{mailto:sc2meisah@gmail.com}{sc2meisah@gmail.com}. $^{*}$Corresponding authors: Jian Zhao, \href{mailto:jianzhao@zgci.ac.cn}{jianzhao@zgci.ac.cn}; Mohamed Elhoseiny, \href{mailto:mohamed.elhoseiny@kaust.edu.sa}{mohamed.elhoseiny@kaust.edu.sa}.}
  \fancyfoot[R]{\footerfont\thepage}
  
  }
\newcommand{\dataset}{EWM}
\newcommand{\cmark}{\ding{51}}
\newcommand{\pmark}{\ding{109}}
\newcommand{\sctwo}{StarCraft~II}
\newcommand{\recordbox}[3][\scriptsize]{%
  \begin{minipage}[t]{0.485\linewidth}
  \VerbatimInput[fontsize=#1,breaklines=true,breakanywhere=true,
    frame=single,framesep=4pt,rulecolor=\color{black!25},
    label={\tiny\sffamily #2}]{#3}
  \end{minipage}}
\title{Esports World Model Dataset}
\author[$\dagger$]{Weiyu Ma}
\author[1]{Zeyuan Gong}
\author[1]{Xuhui Liu}
\author[1]{Wenxuan Zhang}
\author[1]{Jian Ding}
\author[2]{Xinyu Cui}
\author[3,*]{Jian Zhao}
\author[1,*]{Mohamed~Elhoseiny}
\affil[1]{King Abdullah University of Science and Technology (KAUST)}
\affil[2]{Institute of Automation, Chinese Academy of Sciences}
\affil[3]{Beijing Zhongguancun Academy \& Zhongguancun Institute of Artificial Intelligence}
\date{}
\begin{document}

\begin{abstract}
Interactive world models must capture how multiple participants jointly
change a shared environment while observing different parts of it.
Competitive games make this challenge concrete: players cooperate
and compete, while fog of war, team vision and camera placement
determine what each can see. Modeling these interactions requires
data that connect individual observations and actions to shared
state and events.
We introduce the Esports World Model Dataset (\dataset{}), a collection
of human gameplay spanning \sctwo{}, Dota~2 and Counter-Strike~2,
with 3,053 games and maps and approximately 4,880 player-view hours,
including estimated hours for the ongoing CS2 extension. It includes
4,628 completed \sctwo{} views, all ten perspectives for 303 Dota~2
matches, and up to ten views per round in CS2. Across these views,
replay-clock alignment connects rendered video with game-native
actions, structured state and events from the same match.
The three games expose complementary observation conditions:
private fog of war, shared team vision and first-person viewpoints.
In \sctwo{}, player-specific entity records further distinguish live
observations from remembered snapshots and link co-observed entities
across opposing views. Together with temporal alignment, these records
allow researchers to distinguish legitimate information differences
from disagreement about shared state. By connecting human decisions, shared dynamics and player-specific
information, \dataset{} supports research toward jointly controlled
multiplayer world models and interactive environments populated
by teammates and opponents learned from human play.

\end{abstract}


\maketitle
\begin{center}\small
\href{https://www.modelscope.cn/datasets/meisah111/Esports_world_model_datasets}{\faDatabase\enspace Dataset}
\qquad
\href{https://github.com/sc2musa/esports-worldmodel-datasets}{\faIcon{file-alt}\enspace Documentation}
\qquad
\href{https://github.com/vast7777/cs2-demo-to-dataset}{\faGithub\enspace Code}
\end{center}
\input{sections/original_visuals/figure_01}
\input{sections/new_visuals/wall}
\input{sections/introduction}
\input{sections/related_work}
\input{sections/dataset}

\input{sections/validation}

\input{sections/discussion}
\FloatBarrier
\bibliographystyle{abbrvnat}
\bibliography{references}
\end{document}

%% file: sections/original_visuals/figure_01.tex
\begin{figure}[!htbp]
\centering
\includegraphics[width=\linewidth]{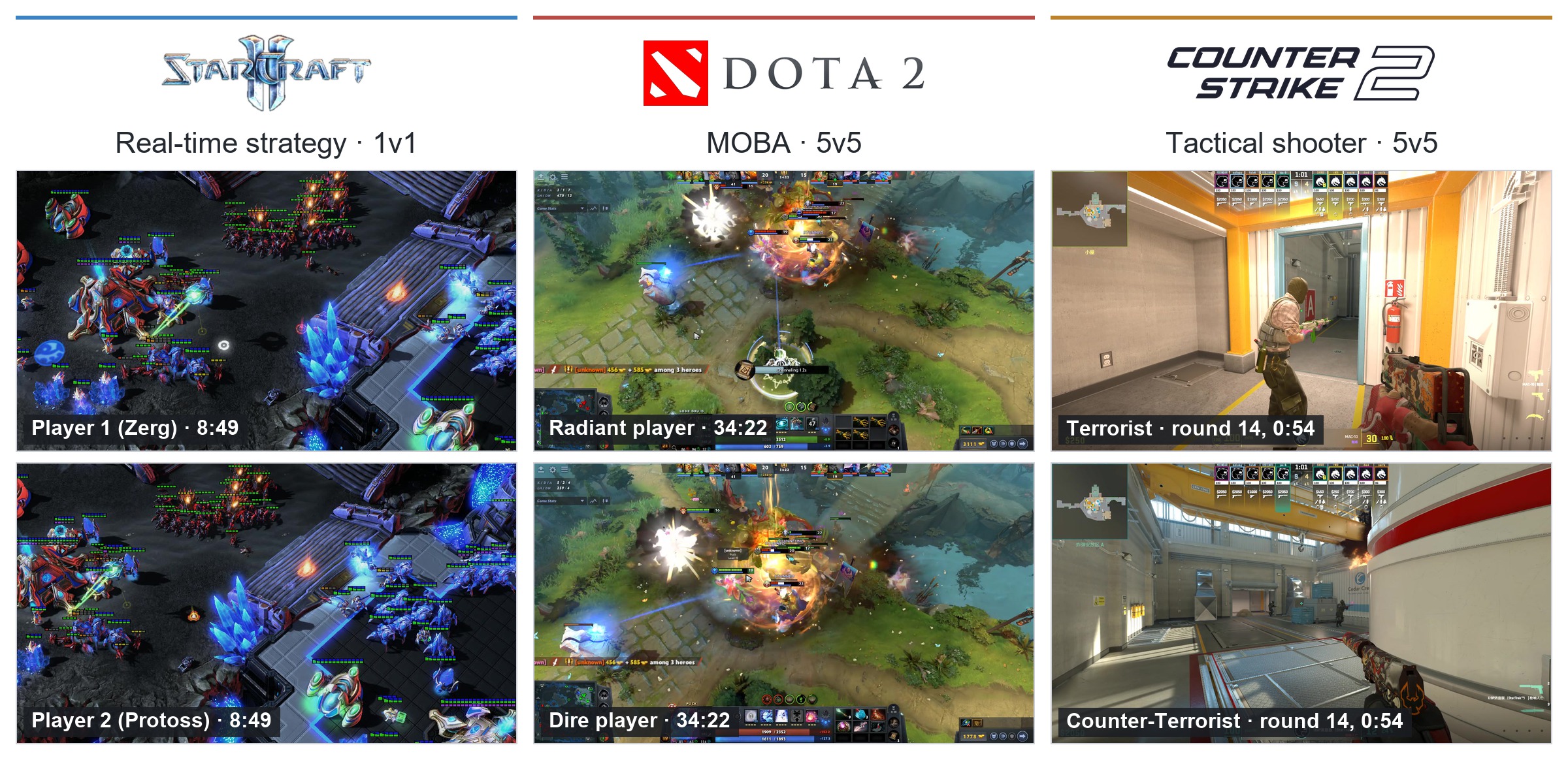}
\caption{\dataset{} records every player's view of real competitive matches. Each column shows two players of one match at the same instant, at the native $1280\times720$ resolution: both players of a \sctwo{} game (Esports World Cup 2025), a Radiant and a Dire player during a Dota~2 team fight (Esports World Cup 2026), and a Terrorist and a Counter-Terrorist in a Counter-Strike~2 round (IEM Krak\'ow 2026). Each view is paired with the player's actions and the engine state on the replay clock of the match.}
\label{fig:cover}
\end{figure}

%% file: sections/new_visuals/wall.tex
\begin{figure}[!t]
\centering
\includegraphics[width=\linewidth]{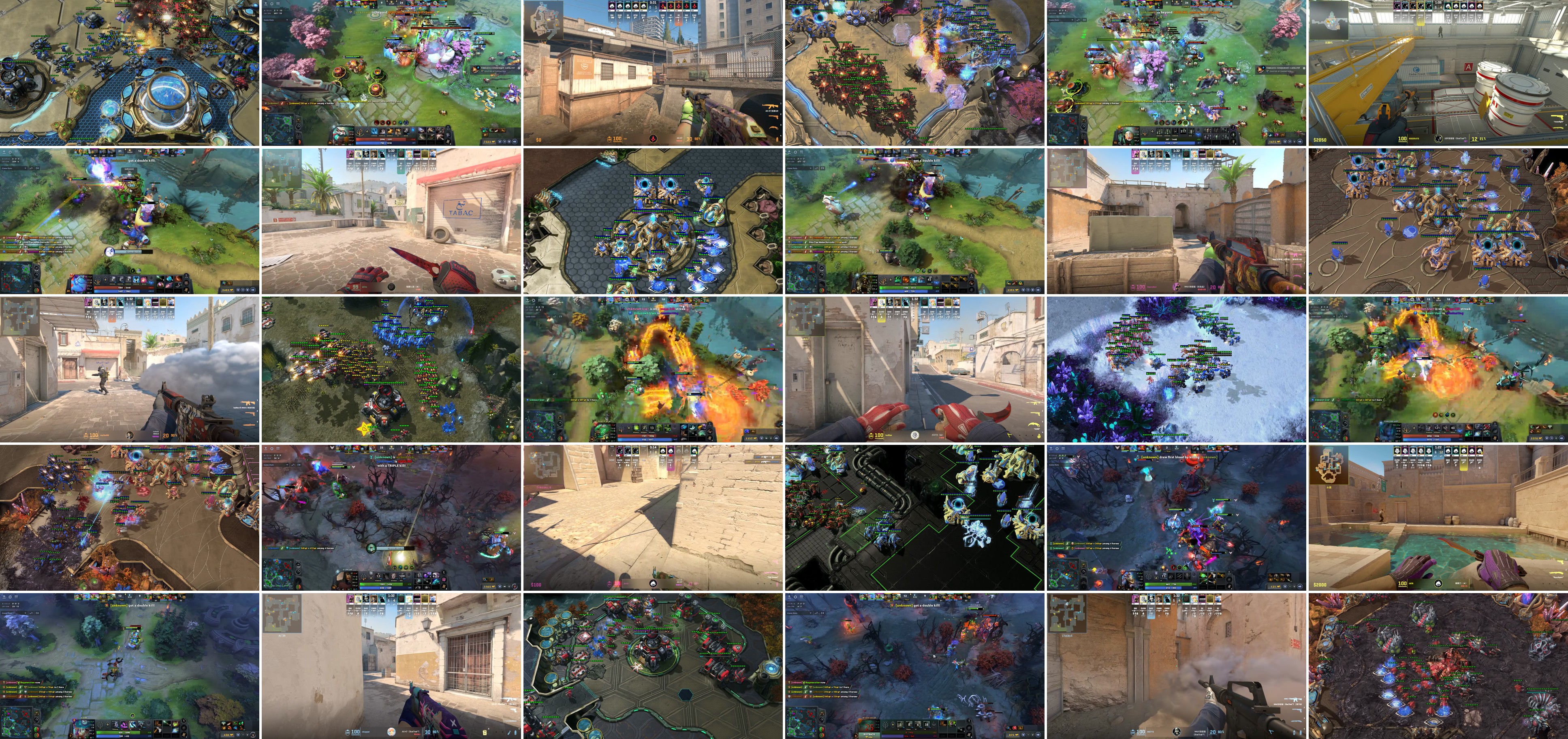}
\caption{Thirty player views from \dataset{}, ten each from \sctwo{}, Dota~2 and Counter-Strike~2 matches.}
\label{fig:wall}
\end{figure}

%% file: sections/introduction.tex
\section{Introduction}
\label{sec:intro}

World models learn how an environment changes as agents interact with it. Recent advances in video generation and action-conditioned prediction have made it possible to simulate increasingly complex visual environments, from Atari and DOOM to commercial games and physical-world scenarios~\cite{ha2018world,hafner2025dreamerv3,alonso2024diamond,valevski2025gamengen,bruce2024genie,nvidia2025cosmos}. For interactive models, the training data must connect what an agent observes, the actions it takes and the subsequent changes in its environment. Multiplayer settings extend this requirement: several agents act in the same world, but each may observe a different part of it.

This extension concerns who changes the world as well as who observes it. In competitive human play, each participant pursues objectives, selects actions and adapts to others; their decisions jointly shape the trajectory. A single-player prediction interface can absorb other actors into the environment, even when those actors behave actively. To study their interaction explicitly, data should associate each participant's actions with their observations and the subsequent shared outcome. Recent multiplayer world models already explore joint action conditioning~\cite{mira2026,mass2026}. A broader direction is to learn worlds populated by \textbf{multiple active participants}, with human gameplay providing examples of both cooperation and competition.

Competitive games provide a concrete setting for connecting this interaction to observation. In \sctwo{}, two opponents can watch the same battle and observe the same units, positions and damage. An army elsewhere on the map may be visible only to its owner, while an enemy building seen earlier remains as a remembered snapshot under fog of war. Both opponents can nevertheless issue commands that change what happens next, including outside the other's current view. Dota~2 players share vision with their teammates but control separate cameras; Counter-Strike~2 players experience the same round through different first-person views. The paired views in \cref{fig:cover} illustrate these differences. A multiplayer world model must preserve \textbf{a common underlying state} while producing observations that reflect \textbf{each player's information}.

Studying this behavior requires more than collecting several videos of the same match. When two views differ, the data should allow a researcher to determine whether they refer to the same replay time and entity, and whether that entity is currently observable to both players. These distinctions are difficult to recover reliably from pixels alone. Structured replay datasets provide actions and state for strategy games~\cite{vinyals2017sc2le,mathieu2023alphastarunplugged}, while recent Counter-Strike datasets pair synchronized player videos with state and action trajectories~\cite{wang2025xego,guo2026egocs}. Dota2-Vis further studies visibility from team-perspective video~\cite{carvalho2026dota2vis}. Together, these resources motivate a more direct connection between rendered gameplay and the information available to individual players.

We introduce the Esports World Model Dataset (\dataset{}) to provide this connection across three competitive game genres. Starting from human match replays, we use the game clients to render player views and extract game-native actions, structured state and events. Replay time associates these outputs within each match. The resulting collection includes \sctwo{}, Dota~2 and Counter-Strike~2, covering private fog of war, shared team vision and first-person viewpoints. This choice provides complementary observation regimes while retaining the native controls and state representation of each game. The recorded matches involve two human competitors in \sctwo{} and ten in Dota~2 and Counter-Strike~2; video availability varies with the collection and, in CS2, player survival. These are trajectories of several decision-makers acting in one environment, with player views exposing different parts of that interaction.

The most detailed observation supervision is available in \sctwo{}. Its player-specific records identify units that are currently observable and distinguish them from remembered snapshots. Live units observed by both players can be associated through their engine identifiers and compared on common state fields. The paired example in \cref{fig:sample_sc2} shows both kinds of information: shared units whose state agrees across players, and hidden units that appear only in their owner's observation. These records connect a visible scene to the player's information state, while keeping engine visibility distinct from whether an entity falls inside the camera window. Dota~2 and Counter-Strike~2 complement this supervision with multi-player video, replay-state streams and event-level records.

This organization motivates two complementary directions for future world models. One is a shared simulation driven by several participants' action streams, enabling them to interact in the same evolving world. The other retains a single focal player while using policies learned from human trajectories to control teammates or opponents. In that setting, policies choose the other participants' actions and a world model predicts their joint consequences. Both directions build on the same need to connect human decisions, shared dynamics and player-specific information, alongside the observation-prediction and hidden-state tasks described later. Complete ten-player examples in \cref{fig:team_dota,fig:team_cs2} make this relationship concrete: the participants experience one event through distinct views, with their actions and state associated on the match timeline.

\begin{samepage}
Our contributions are threefold:
\begin{itemize}[leftmargin=*,itemsep=3pt,topsep=3pt]
  \item We release \textbf{a multiplayer, multi-genre dataset of human interaction}, pairing rendered player views with game-native actions, state and events. Private fog of war, shared team vision and first-person viewpoints provide complementary settings for studying how several participants jointly change a world under different information constraints (\cref{sec:dataset}).
  \item We provide \textbf{\sctwo{} player-observation records} that distinguish live visibility from remembered snapshots and associate co-observed entities across opposing views. This supervision makes it possible to separate legitimate information differences from disagreement about shared state (\cref{sec:visibility}).
  \item We document \textbf{replay-clock alignment and initial consistency checks}, with game-specific joining rules and linked visual, action and state examples. These provide a reproducible basis for constructing cross-player samples and inspecting their timing and observation semantics (\cref{sec:alignment,sec:quality}).
\end{itemize}
\end{samepage}

We organize the report around the sources, extraction procedures and semantics of the released data, followed by coverage statistics and initial quality checks. The final sections describe research uses, evaluation requirements and future directions.

%% file: sections/related_work.tex
\section{Related Work}
\label{sec:related}

\paragraph{World models and multiplayer consistency.}
Single-observer world models learn dynamics for control or generate interactive gameplay~\cite{ha2018world,hafner2025dreamerv3,alonso2024diamond,valevski2025gamengen,bruce2024genie}. Multiplayer systems increasingly generate several views jointly or maintain a shared representation~\cite{solaris2026,multiworld2026,mira2026,gammaworld2026,khora2026}. In particular, MIRA conditions on multiple players' actions using bot-collected Rocket League trajectories~\cite{mira2026}. COMBO reasons from partial views in cooperative environments~\cite{zhang2025combo}, while MASS, WorldWeaver and ConsistWorld study explicit state, memory or evidence sharing~\cite{mass2026,worldweaver2026,consistworld2026}. These methods motivate datasets that associate multiple observations and action streams with a shared environment. \dataset{} contributes human competitive trajectories across three games and adds \sctwo{} records that identify which live state is available to each player.

\paragraph{Video and embodied interaction data.}
HowTo100M pairs instructional video with transcribed narration~\cite{miech2019howto100m}, while InternVid builds video--text pairs using generated descriptions~\cite{wang2024internvid}. For learning actions from online gameplay, VPT uses an inverse dynamics model trained on a smaller labeled collection to annotate previously unlabeled video~\cite{baker2022vpt}. Robotic resources provide a complementary approach: DROID records manipulation demonstrations across diverse scenes~\cite{khazatsky2024droid}, and Open X-Embodiment standardizes datasets from multiple robot platforms~\cite{openx2023}. These efforts connect observations to language or actions at scale. \dataset{} focuses on game-native commands, replay state and player-observation semantics; its three games retain their own action and state representations.

\paragraph{Rendered gameplay and replay datasets.}
The closest Counter-Strike resources pair multiple player views with structured trajectories. X-Ego-CS includes synchronized perspectives from professional matches and a teammate/opponent location-prediction evaluation~\cite{wang2025xego}; EgoCS-400K provides a substantially larger Counter-Strike collection with action, state and language annotations~\cite{guo2026egocs}. Dota2-Vis provides dual-team video and minimap annotations for visibility analysis~\cite{carvalho2026dota2vis}. \Cref{tab:related} places these approaches within the broader gameplay-data landscape. \dataset{} extends this coverage to complementary strategy-game observation regimes, with player-specific entity records in \sctwo{}.

\input{sections/original_visuals/table_01}

\paragraph{Partial observation and hidden-state estimation.}
Brood War and \sctwo{} datasets have long supported learning from structured replays~\cite{lin2017stardata,wu2017msc,vinyals2017sc2le,mathieu2023alphastarunplugged}; Defogger predicts hidden strategy-game state~\cite{synnaeve2018defogger}. \dataset{} connects that problem to rendered multi-view gameplay and distinguishes observation labels from privileged state. This distinction is also complementary to synchronized real-world egocentric video, where a game engine is unavailable as a source of entity records~\cite{grauman2022ego4d,grauman2024egoexo4d}.

%% file: sections/original_visuals/table_01.tex
\begin{table}[!htbp]
\centering
\caption{Comparison with gameplay datasets. Human: recorded from human players. Exact actions: actions logged by the game rather than inferred from video. Engine state / events: game state or events from the engine. Multi-POV: synchronized views of several players in one match. Cross-genre: several game genres. \cmark{} yes, \pmark{} partial, -- no.}
\label{tab:related}
\footnotesize
\renewcommand{\arraystretch}{1.2}
\setlength{\tabcolsep}{4pt}
\begin{tabularx}{\linewidth}{@{}l>{\raggedright\arraybackslash}X*{5}{>{\centering\arraybackslash}p{1.04cm}}@{}}
\toprule
Dataset & Game, scale and data source & \makecell[b]{Human} & \makecell[b]{Exact\\actions} & \makecell[b]{Engine\\state /\\events} & \makecell[b]{Multi-\\POV} & \makecell[b]{Cross-\\genre} \\
\midrule
\rowcolor{blue!8}\textbf{\dataset{} (ours)} & \sctwo{}, Dota~2 and CS2; 3,053 games and maps and about 4,880 player-view hours from human ladder and professional replays & \cmark & \cmark & \cmark & \cmark & \cmark \\
EgoCS-400K~\cite{guo2026egocs} & CS/CS2; 10,000+ hours and 400K+ first-person videos from professional match replays & \cmark & \cmark & \cmark & \cmark & -- \\
MIRA~\cite{mira2026} & Rocket League; 10,000 match-hours of bot-generated 2v2 gameplay with four synchronized views & -- & \cmark & \cmark & \cmark & -- \\
MultiWorld~\cite{multiworld2026} & It Takes Two; 100+ hours of paired human gameplay with synchronized controls and views & \cmark & -- & \cmark & \cmark & -- \\
Dota2-Vis~\cite{carvalho2026dota2vis} & Dota~2; 144 professional matches recorded from two team perspectives & \cmark & -- & \pmark & \pmark & -- \\
DIAMOND / CS:GO~\cite{pearce2021csgo,alonso2024diamond} & CS:GO; 95 hours of human Deathmatch gameplay, used by DIAMOND & \cmark & \cmark & \pmark & -- & -- \\
MineRL~\cite{guss2019minerl} & Minecraft; 60M+ human state--action pairs across multiple tasks & \cmark & \cmark & \cmark & -- & -- \\
Matrix-Game-MC~\cite{zhang2025matrixgame} & Minecraft; 2,700+ hours of video and 1,000+ hours of action-labelled clips & \pmark & \cmark & -- & -- & -- \\
NitroGen~\cite{magne2026nitrogen} & 1,000+ games; 40,000 hours of public gameplay with actions inferred from controller overlays & \cmark & \pmark & -- & -- & \cmark \\
\bottomrule
\end{tabularx}
\end{table}

%% file: sections/dataset.tex
\section{The \dataset{} Dataset}
\label{sec:dataset}

\dataset{} is organized around a match and its replay timeline. Each \emph{player view} is rendered from one participant's perspective and associated with actions, state and events through the game's replay clock. The three games share this organization but differ in their observation rules, native controls and available structured records. We first describe the sources and extraction procedures, then use concrete examples to explain how the modalities should be read together.

\subsection{Games, sources and modalities}
\label{sec:games}

The collection combines high-ranked ladder and professional play (\cref{fig:sources}). In \sctwo{}, most games are high-ranked ladder matches; 88\% of views with known rating are at least 5,000 MMR. The 190 professional games come from the Esports World Cup 2025, Kung Fu Cup 2026 and Rongyi Cup Season~3 (2026), on game versions 5.0.14 and 5.0.15. Dota~2 sources are The International 2025, Esports World Cup 2026 and regional qualifiers of the Road to ENC 2026. The CS2 core combines 30 professional and 20 matchmaking maps, while the extended collection contains additional professional series. \Cref{tab:events} gives the event-level counts and identifies sources that remain unlabelled.

\input{sections/original_visuals/figure_02}

The games provide different settings for multi-view modeling. \sctwo{} pairs two opponents whose armies explore a shared map under private fog of war. Dota~2 provides ten player cameras with shared vision within each team. Counter-Strike~2 provides first-person views of short rounds in which the available views decrease as players die. These differences determine both what a model can observe and how a prediction should be evaluated.

\input{sections/original_visuals/table_02}

\Cref{tab:modalities} summarizes the native streams. They are associated by replay time, but the action spaces are game-specific. The CS2 core records button states and view angles; recovery of raw mouse movement is a feature of the extended pipeline. Likewise, the core uses a round-level frame/tick rule, while the extended pipeline supplies explicit per-frame alignment tables. The timing conventions are detailed below.

\subsection{From replays to paired observations}
\label{sec:construction}
\label{sec:pipelines}

A replay stores information used by the game engine to reproduce a recorded match. We use the Windows game clients to render player views and associate them with structured records on the replay timeline. \Cref{fig:pipeline} summarizes the common organization of these processes: obtain player views and native records, associate them by replay time, and package the paired outputs. The game-specific implementations differ in how they obtain observations, decode state and map frames to the replay clock.

\paragraph{\sctwo{}.}
The replay is loaded in its matching game build and replayed once per player with fog of war enabled. At each game loop, the extraction records the camera image, minimap and player-specific observation, together with actions and camera metadata. Matching the build allows both views to traverse the same replay timeline and retain shared tags for live units.

\paragraph{Dota~2.}
The replay is parsed into entity-property changes and game messages, including combat logs and orders attributed to their issuing players. Persistent clients render the ten player slots in Player View mode in 120-second segments. The per-frame alignment records associate each independently rendered video with replay ticks and with actions in its frame intervals.

\paragraph{Counter-Strike~2.}
The parser extracts round boundaries, player states, inputs and events. A recording process follows each player in first-person spectator mode, with the playback controls hidden. Clips run from round start until two seconds after the player's death or the end of the round. The extended pipeline also checks client-build compatibility and recorded-player identity.

\input{sections/unified_pipeline}

\subsection{What a player view contains}
\label{sec:player_views}

The montage in \cref{fig:wall} illustrates the visual coverage of the collection. To understand what these videos provide to a model, it is useful to first inspect the individual observation interfaces (\cref{fig:single_sc2,fig:single_dota2,fig:single_cs2}), then compare players at a common replay time. The montage samples different matches; the paired and team-view examples below establish the within-match correspondence.

In \sctwo{}, a player controls an economy and an army while the camera exposes a local region of the map. The camera video omits the interface, and the minimap is a separate stream. A player's raw entity observation can contain units outside this camera window, which is why camera RGB and engine-visible state must be treated as distinct modalities.

\input{sections/new_visuals/single_sc2}

Dota~2's Player View includes the player's camera, interface and minimap. Teammates share vision but can attend to different parts of the same encounter. The first-person CS2 view includes its HUD and radar and is tied to one player's position and viewing direction. These differences affect the information budget of a prediction or policy-learning task even when all three games supply video, actions and state.

\input{sections/new_visuals/single_dota}
\input{sections/new_visuals/single_cs2}

\subsection{Observation semantics and paired examples}
\label{sec:visibility}
\label{sec:samples}

The examples in \cref{fig:sample_sc2,fig:sample_dota,fig:sample_cs2} connect the rendered views to records from the same encounter. Reading the modalities together requires separating the player's observations from privileged replay state, and separating replay time from local frame indices.

\paragraph{\sctwo{}: current observations and remembered state.}
The released raw observation is player-specific and camera-independent. It records the player's own units, neutral units and enemy units exposed by the engine's observation rules. An engine-visible enemy can therefore be outside the current RGB frame. An enemy record marked \emph{visible} describes a live observation. When an observed enemy structure returns under fog, the engine may retain a \emph{snapshot} of its last-seen type and position. This memory record carries no current health or orders and may persist after the structure has disappeared. Enemy units that are neither currently observed nor retained as snapshots are absent.

\Cref{fig:sample_sc2} shows this distinction in a single encounter. Units A and B are observed by both players, with matching common state, while only the owner has their orders. The units marked C appear in the owner's frame and are absent from the opponent's observation.

\input{sections/original_visuals/figure_06}

\Needspace{12\baselineskip}
For a precise cross-view join, let $L_i(t)$ denote the engine tags of live units marked visible in player $i$'s observation at replay time $t$. The set of live co-observed entities for players $i$ and $j$ is
\begin{equation}
  C_{ij}(t) = L_i(t) \cap L_j(t).
  \label{eq:coobserved}
\end{equation}
The join is scoped to one match and replay time. On $C_{ij}(t)$, the audit in \cref{sec:quality} compares common fields such as type, position, health and shield. Owner-only orders are outside this comparison. Remembered snapshots are excluded from the live-state join and examined separately through their last-seen information.

\textbf{Engine visibility, camera-window inclusion and rendered pixels are different quantities.} The structured records identify the player's observation state; camera metadata helps locate that observation in the view. Evaluating generated pixels additionally requires a validated association between image content and entities, including camera projection and rendering effects.

\paragraph{Dota~2: shared vision and independent cameras.}
Five teammates share vision but move their cameras independently. The rendered Player View includes the interface and minimap, while the parsed state stream contains information beyond a single player's view. Per-entity player-visibility labels are not part of the current release. At the kill event in \cref{fig:sample_dota}, the victim and killer reach replay tick 85,916 at different local frame indices. Their alignment records establish the correspondence between the views.

\input{sections/original_visuals/figure_07}

The ten-player expansion in \cref{fig:team_dota} places the same kill in its team context. Five Radiant views and five Dire views surround the victim and killer highlighted in the pair. Teammates can look at different regions despite sharing team vision, so visual differences within a team need not imply inconsistent world state. Across teams, the observation rules introduce an additional information boundary. This example motivates evaluating camera agreement and player information separately, using the event tick to associate all participants.

\input{sections/new_visuals/team_dota}

\paragraph{Counter-Strike~2: frame intervals and partial-round views.}
The sample in \cref{fig:sample_cs2} uses frame 952 of round~14. This frame covers ticks 132,354--132,357, and the displayed state is sampled at tick 132,356. The kill occurs 16 ticks, or 0.25 seconds, after that state sample. Because clips end shortly after each player's death, the set of available views changes during a round. The release supplies player state and events, but no entity-visibility labels.

\input{sections/original_visuals/figure_08}

\Cref{fig:team_cs2} adds all ten POVs at an earlier instant of the same round, just before its first kill. At tick 132,100, FalleN is about 264 game units above donk and looks steeply downward; their views therefore describe the same encounter from substantially different positions and directions. The event at tick 132,117 identifies the killer and victim without requiring the other eight views to show the encounter. The earlier ten-view example and the later pair serve different purposes: one preserves the full participant context, while the other exposes a local interaction in more detail. Their separate records are shown in \cref{fig:records_cs2_team,fig:records_cs2}.

\input{sections/new_visuals/team_cs2}

\FloatBarrier
\subsection{Gameplay coverage and temporal examples}
\label{sec:gallery}

The galleries show how the rendered observations vary across maps and matches. The accompanying temporal examples follow a single game, fight or round and associate the displayed frames with event or trajectory rows. Together, they connect visual coverage to the temporal structure of the released sequences.

\paragraph{\sctwo{}.}
\Cref{fig:gallery_sc2} presents a range of map and player views. \Cref{fig:log_sc2} follows both players through one game, connecting changes in the scene to resources, army supply, recorded actions and accumulated losses.

\input{sections/original_visuals/figure_09}

\Cref{fig:gallery_sc2_pairs} complements the individual-view gallery with both opponents at the same game loop on each of six maps. The 55\%-of-game sampling point is shared within a pair, rather than being a common absolute game loop across matches. The pairs illustrate how camera placement and fog of war expose different regions of one evolving match.

\input{sections/new_visuals/gallery_sc2_pairs}
\input{sections/original_visuals/figure_10}

\paragraph{Dota~2.}
\Cref{fig:gallery_dota} samples fights across professional matches. \Cref{fig:log_dota} follows one team fight through successive kills, with messages and combat-log fields supplying event times and outcomes.

\input{sections/original_visuals/figure_11}

The complete-team gallery in \cref{fig:gallery_dota_team} expands this view to two further professional matches. Keeping all player slots makes it possible to connect the focal encounter to the surrounding participants instead of selecting only the killer's camera. The highlighted killer and victim establish an event anchor; shared team vision does not require all teammates to place their cameras on that anchor.

\input{sections/new_visuals/gallery_dota_team}
\input{sections/original_visuals/figure_12}

\input{sections/new_visuals/log_dota_team}

\Cref{fig:log_dota_team} follows the same fight with all ten views at three replay ticks. The five-kill event log in \cref{fig:log_dota} provides the sequence of outcomes, while the team panels retain the participants' changing visual context. Team damage is measured over the preceding three seconds, and the killer's orders over the preceding two seconds. Together, the records and panels offer an example for studying how local decisions accompany a shared interaction over time.

\paragraph{Counter-Strike~2.}
\Cref{fig:gallery_cs2} shows different maps and encounters. \Cref{fig:log_cs2} follows a round through its kill events and links those events to per-tick player state.

\input{sections/original_visuals/figure_13}

\Cref{fig:gallery_cs2_team} includes ten POVs from rounds on two further maps, sampled one second before the first kill. This event-relative sampling preserves views before player elimination starts to reduce coverage. The panels show the different positions and viewing directions from which participants enter an encounter.

\input{sections/new_visuals/gallery_cs2_team}
\input{sections/original_visuals/figure_14}

The full-round panels in \cref{fig:log_cs2_team} make the changing view availability explicit. Grey tiles mark eliminated players at the selected instants, consistent with clips ending two seconds after death. Joint prediction samples can use a per-player availability mask to account for these variable clip lengths. The original event table links the sequence to kill times and the killer's health.

\input{sections/new_visuals/log_cs2_team}

\FloatBarrier
\subsection{Timing, file organization and released records}
\label{sec:alignment}
\label{sec:records}

We distinguish replay time, trajectory row index and video frame index. Replay time is an integer game loop in \sctwo{} or a replay tick in Dota~2 and Counter-Strike~2. A frame index is local to one rendered view and must not be treated as a shared clock without checking the alignment records.

\sctwo{} stores one trajectory row and one frame per game loop. Using $n$ for the trajectory row index, the action in row $n$ is assigned to the interval between observations $n$ and $n+1$; the explicit game-loop field provides replay time. Dota~2 records each frame's replay tick and orders issued in the frame interval. Cross-view joins must account for repeated or missing samples.

In the core CS2 set, frame $k$ covers ticks $t_r+4k$ through $t_r+4k+3$, where $t_r$ is the recorded round-start tick. This identifies an interval rather than a single engine state. The extended pipeline records frame-to-tick alignment explicitly. Views should be aligned by replay time and available ticks rather than by their endpoints.

\paragraph{\sctwo{} records.}
\label{sec:sc2_records}
Each player view has a Parquet trajectory, camera and minimap videos, and metadata for games, views, checksums and per-build unit/ability vocabularies. \Cref{fig:records_sc2} expands the action, timing and unit records associated with the earlier paired example. The trajectory row index and explicit game loop serve different roles.

\input{sections/original_visuals/figure_15}

\paragraph{Dota~2 records.}
\label{sec:dota_records}
Per-match entity-change and message tables are accompanied by per-slot video, alignment and action tables. \Cref{fig:records_dota} preserves the separate replay-tick, video-time and game-clock fields at the shared event. Storage mirrors may contain overlapping match IDs; these must be deduplicated before counting or splitting.

\input{sections/original_visuals/figure_16}

\paragraph{Counter-Strike~2 records.}
\label{sec:cs2_records}
Map-level metadata and event streams are accompanied by per-round, per-player videos and state tables. \Cref{fig:records_cs2} separates the state inside the displayed frame from the later kill event. Experiments should select complete video/state pairs using the release quality reports and record the selected match IDs and release revision.

\input{sections/original_visuals/figure_17}

The additional record pair in \cref{fig:records_cs2_team} belongs to the earlier ten-player snapshot. Frame 888 covers ticks 132,098--132,101; its illustrated state row is at tick 132,100, and the subsequent kill is at tick 132,117. The 17-tick separation from the sampled state to the event is about 0.27~s at 64 ticks/s. The state and event rows link the pre-kill observation to the subsequent outcome.

\input{sections/new_visuals/records_cs2_team}

\FloatBarrier
\subsection{Release coverage and statistics}
\label{sec:coverage}

\Cref{tab:stats} gives the scale of the collection: \textbf{3,053 games and maps and about 4,880 player-view hours}, including the CS2 extension. \emph{Timeline hours} count each match once; \emph{player-view hours} sum over player views. Dota~2 video durations include the pre-game phase. Structured-data units and the extension estimates are specified in the table.

\input{sections/original_visuals/table_03}

Of the 2,320 indexed SC2 games, 2,314 are training-ready and provide 4,628 completed player views. The train/validation/test split contains 2,088/117/109 games, with both views assigned together. The remaining six indexed games are not counted as training-ready. Median game duration is 10.3 minutes, with a 10th--90th percentile range of 4.7--18.5 minutes. \Cref{fig:sc2dist} shows the duration, matchup and rating distributions. Among 4,052 ladder views with known rating, median MMR is 5,464 (10th--90th percentile: 4,913--5,737); 88\% are at least 5,000 and 44\% at least 5,500. These are view-level statistics and tournament games carry no MMR. The rating distribution characterizes the high-ranked ladder component. The release occupies 144~GB of trajectories and 4.9~TB of video.

Dota~2 contains 303 professional matches involving 381 players from 82 teams and 8.4B entity-state changes. The CS2 core contains 30 professional maps covering 17 teams and 90 players, plus 20 matchmaking maps with 175 players. Neither the Dota~2 replays nor the released CS2 files record player rank or matchmaking rating.

The CS2 extension contains 380 additional professional maps being added to the collection. Its approximately 265 timeline hours, 1,050 view-hours and 570M rows are estimated from core-set averages, as indicated in \cref{tab:stats}. The extended set spans 171 professional series, 61 teams and 327 players. \Cref{tab:events} lists the events, including the IEM Cologne Major 2026 and BLAST Bounty 2026 Season~2. Teams and players are deduplicated within the reported collection totals; per-event counts need not add to those distinct totals. These sources support held-out-series and event-level evaluation splits.

\input{sections/original_visuals/figure_18}
\input{sections/new_visuals/table_events}

%% file: sections/original_visuals/figure_02.tex
\begin{figure}[!htbp]
\centering
\includegraphics[width=\linewidth]{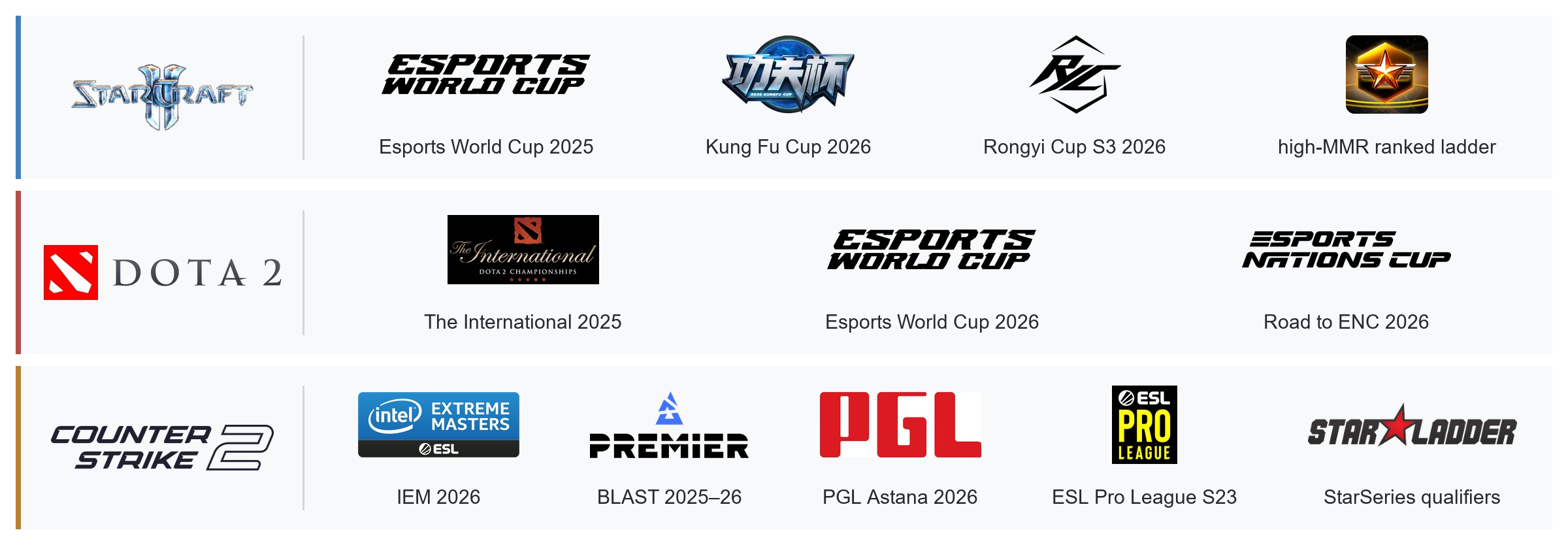}
\caption{Where the matches come from. \sctwo{} combines tournament games with high-MMR ranked games; all Dota~2 matches and the Counter-Strike~2 extended set are professional, from the events shown and their qualifiers.}
\label{fig:sources}
\end{figure}

%% file: sections/original_visuals/table_02.tex
\begin{table}[!htbp]
\centering
\caption{Modalities per game. All streams of a match share one clock.}
\label{tab:modalities}
\small
\begin{tabular}{@{}P{0.13\linewidth}P{0.26\linewidth}P{0.26\linewidth}P{0.26\linewidth}@{}}
\toprule
 & \sctwo{} & Dota~2 & Counter-Strike~2 \\
\midrule
Views & 2 per game & 10 per match & up to 10 per round \\
Video & camera view without interface, 1280$\times$720; minimap, 256$\times$256; 22.4 fps & Player View with interface and minimap, 1280$\times$720, 30 fps & first-person view with interface and radar, 1280$\times$720, 16 fps \\
Actions & unit commands (ability, target, queued flag) and camera moves & unit orders of all ten players, attributed to the issuing player & button states (move, fire, jump, duck, use, reload), view angles, raw mouse movement \\
State & observed units per view (type, owner, alliance, visibility, position, health, shields, energy, orders, buffs), resources, score, upgrades, camera & complete entity state as a lossless change stream & per-tick position, velocity, view angles, health, armor, weapon, money \\
Events & unit deaths; game result & all game messages, including the combat log & kills, damage, bomb plants, round ends \\
Clock & game loop, 22.4/s & replay tick, 30/s & replay tick, 64/s \\
Frame alignment & one frame per game loop & replay tick of every frame & replay tick of every frame \\
\bottomrule
\end{tabular}
\end{table}

%% file: sections/unified_pipeline.tex
\begin{figure}[!htbp]
\centering
\includegraphics[width=\linewidth]{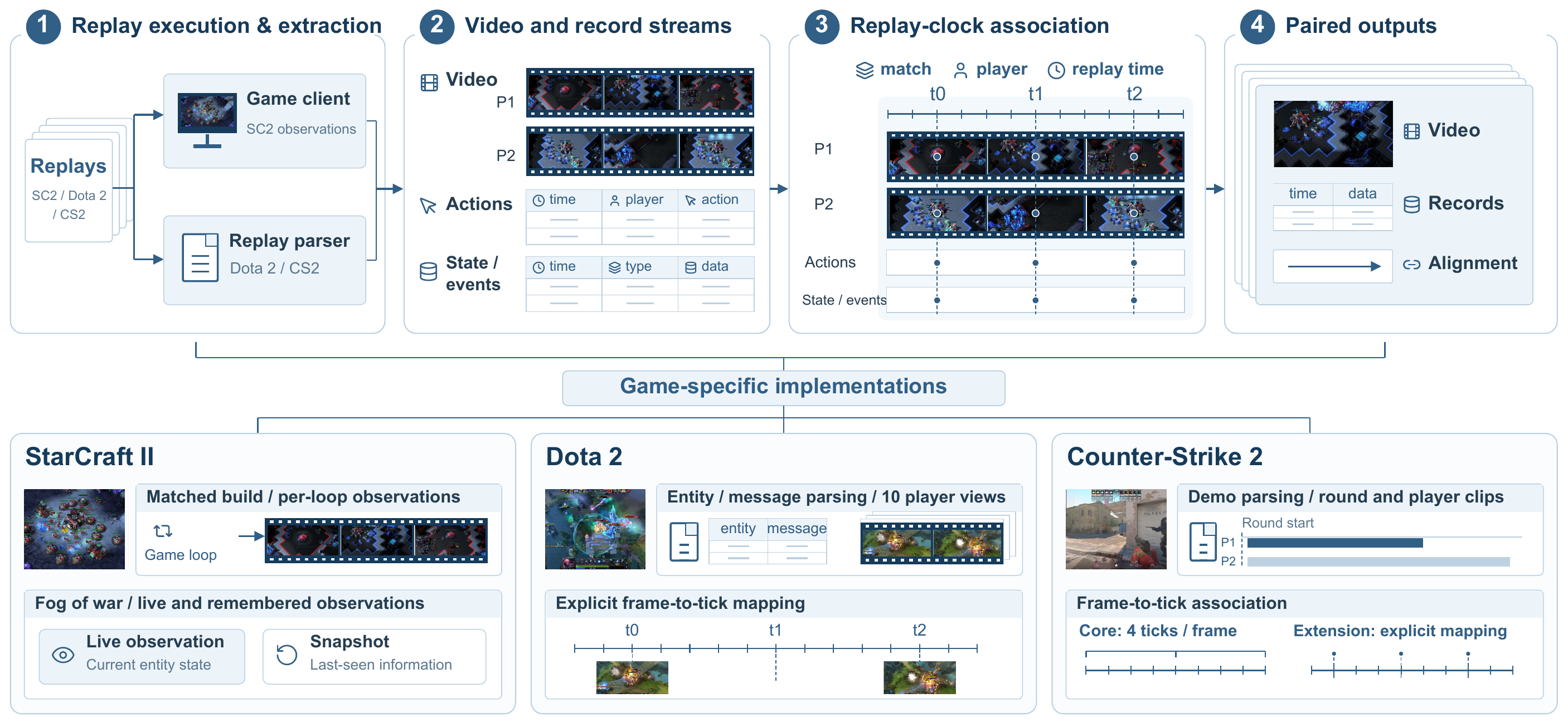}
\caption{Shared data-construction stages and game-specific implementations. Native clients render player views, while game-specific extraction supplies actions and records; SC2 observations come from the replaying engine. Streams are associated on each match's replay clock and packaged as paired outputs. The CS2 core uses four-tick frame intervals; its extension uses explicit mappings. This schematic summarizes separately implemented processes, and its game images and record tables are illustrative.}
\label{fig:pipeline}
\end{figure}

%% file: sections/new_visuals/single_sc2.tex
\begin{figure}[!htbp]
\centering
\includegraphics[width=0.72\linewidth]{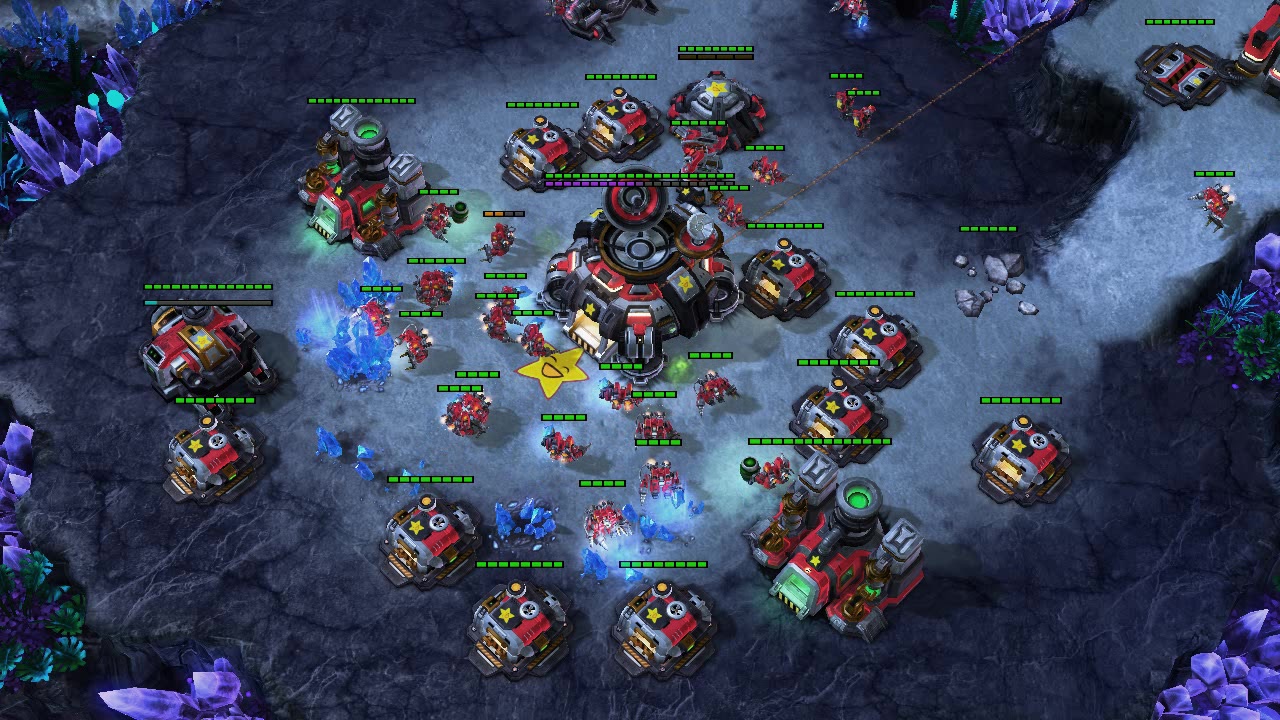}
\caption{One \sctwo{} player view at the native $1280\times720$ resolution (map Winter Madness, Player~2, Terran). The camera covers a small part of the map; the minimap is released as a separate video stream.}
\label{fig:single_sc2}
\end{figure}

%% file: sections/new_visuals/single_dota.tex
\begin{figure}[!htbp]
\centering
\includegraphics[width=0.72\linewidth]{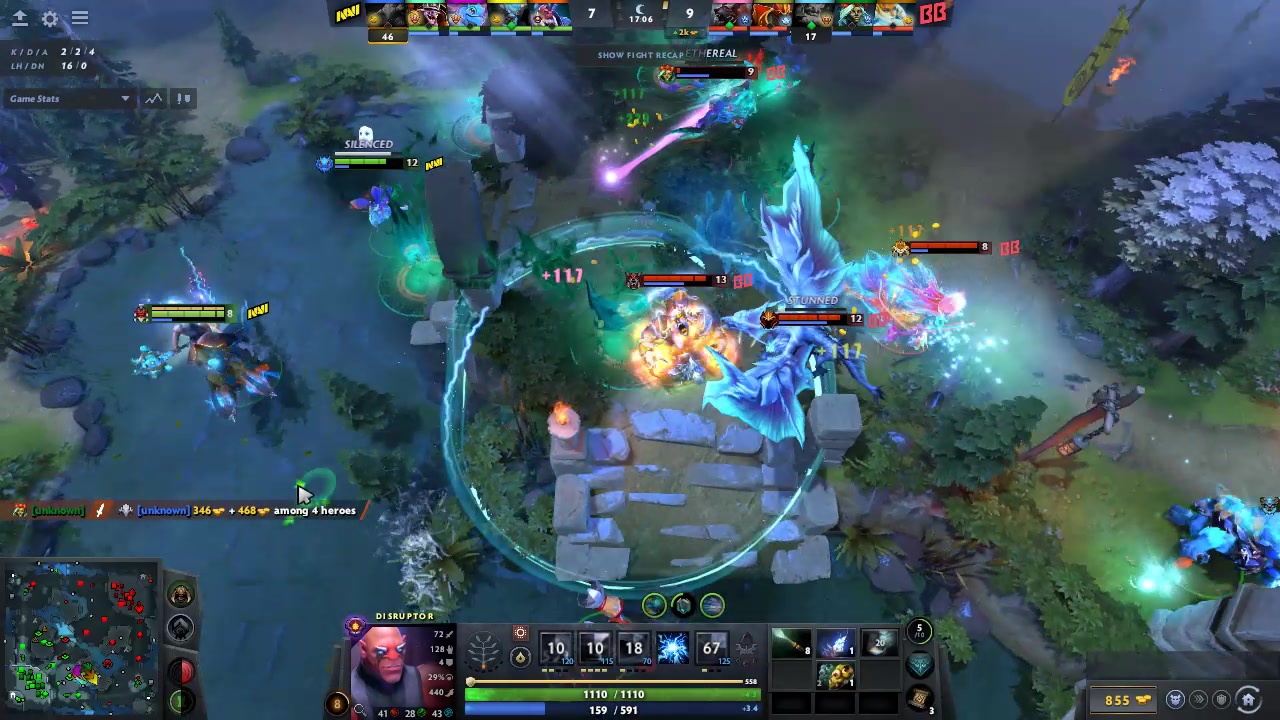}
\caption{One Dota~2 Player View at the native $1280\times720$ resolution (The International 2025, match 8447715678), a third of a second before the player's hero scores a kill. Player View mode shows the player's own camera, interface and minimap.}
\label{fig:single_dota2}
\end{figure}

%% file: sections/new_visuals/single_cs2.tex
\begin{figure}[!htbp]
\centering
\includegraphics[width=0.72\linewidth]{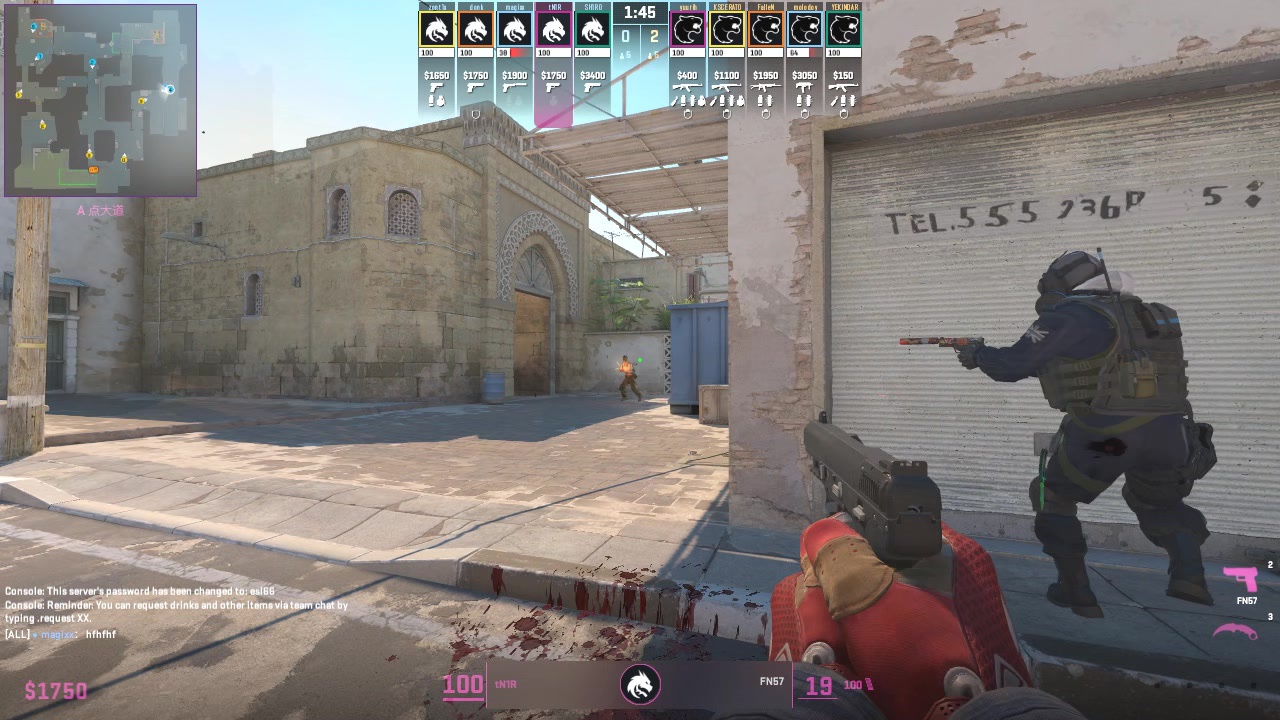}
\caption{One Counter-Strike~2 POV at the native $1280\times720$ resolution (IEM Krak\'ow 2026, Spirit vs.\ FURIA, Dust2, round 3), a third of a second before the player's kill: the first-person view with the player's HUD.}
\label{fig:single_cs2}
\end{figure}

%% file: sections/original_visuals/figure_06.tex
\begin{figure}[!htbp]
\centering
\includegraphics[width=\linewidth]{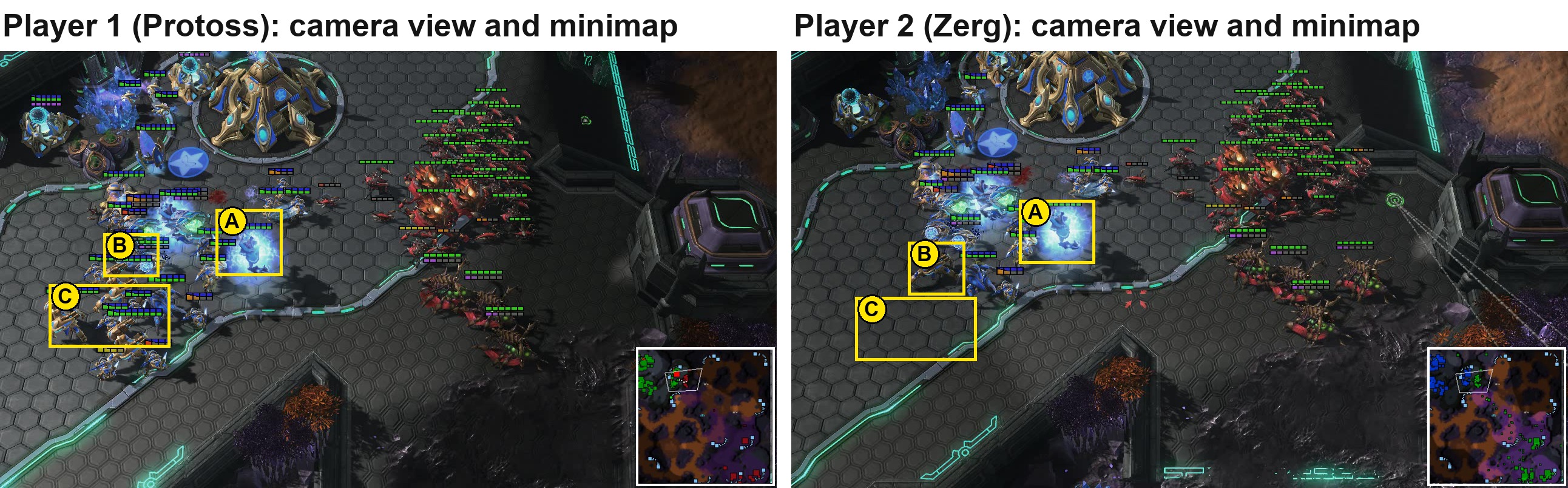}\\[4pt]
\footnotesize
\input{figures/samples/sc2_records.tex}
\caption{\sctwo{}: game loop 12,057 of a Protoss-vs-Zerg game (Kung Fu Cup 2026, map Taito Citadel); both cameras are on the same standoff. A (an Archon) and B (an Immortal) are in both players' observations, with the same tag, position, health and shield; only the owner's record contains their attack orders. C marks two Immortals that Player~1 sees on screen but that are under fog for Player~2: they are absent from Player~2's observation, and the same spot in Player~2's frame is empty.}
\label{fig:sample_sc2}
\end{figure}

%% file: figures/samples/sc2_records.tex
\begin{tabular}{@{}cllllll@{}}
\toprule
 & Unit & Tag & Player 1 (Protoss) & Player 2 (Zerg) & Position & Health/shield \\
\midrule
A & Archon & 4392747010 & own, attacking & enemy, no order & (47.4, 104.3) & 10/350 \\
B & Immortal & 4398252034 & own, attacking & enemy, no order & (42.7, 103.5) & 200/100 \\
C & Immortal $\times$2 & 4394057732, 4407164956 & own, attacking & absent (fog) & -- & 200/100 \\
\bottomrule
\end{tabular}

%% file: sections/original_visuals/figure_07.tex
\begin{figure}[!htbp]
\centering
\includegraphics[width=\linewidth]{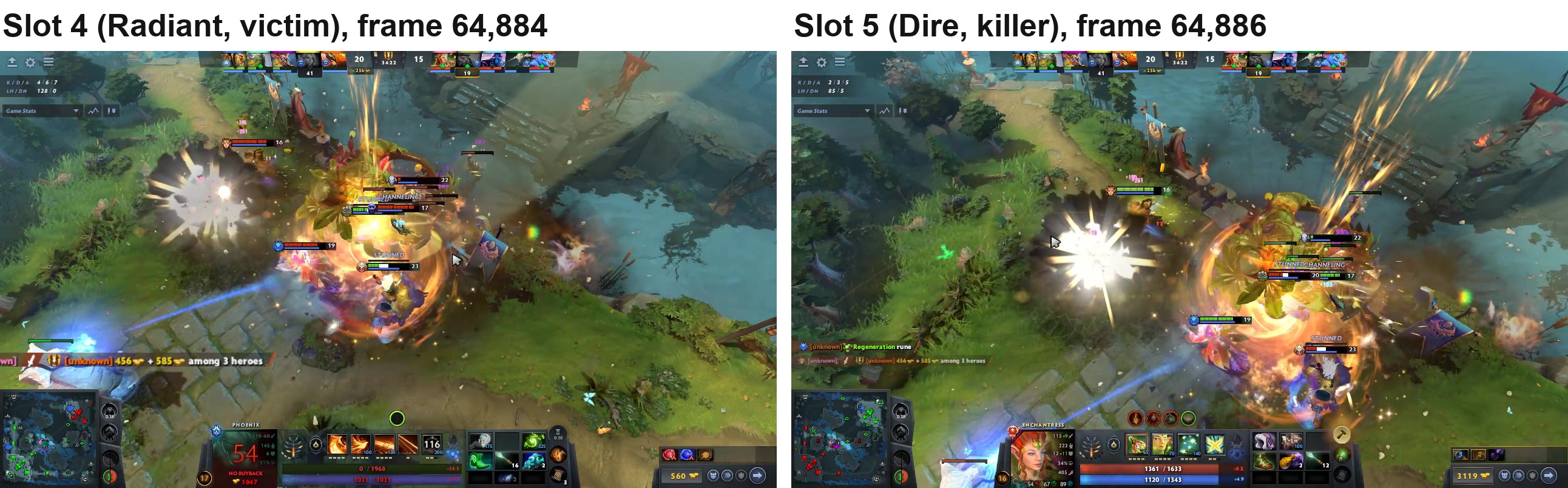}\\[4pt]
\footnotesize
\input{figures/samples/dota_records.tex}
\caption{Dota~2: a hero kill in a team fight (Esports World Cup 2026, match 8893208737). Top: the views of the victim (slot~4) and the killer (slot~5) at the replay tick of the kill. Bottom: the released records of this shared replay tick.}
\label{fig:sample_dota}
\end{figure}

%% file: figures/samples/dota_records.tex
\begin{tabular}{@{}llll@{}}
\toprule
File & Replay tick & Record & Content \\
\midrule
messages & 85916 & chat event & hero kill: slot 5 killed slot 4, bounty 456 gold \\
messages & 85916 & combat log & death of a Radiant hero, attacker team Dire \\
slot04/alignment & 85916 & frame 64,884 & 30 orders by slot 4 in the preceding 2 s \\
slot05/alignment & 85916 & frame 64,886 & 5 orders by slot 5 in the preceding 2 s \\
\bottomrule
\end{tabular}

%% file: sections/new_visuals/team_dota.tex
\begin{figure}[!htbp]
\centering
\includegraphics[width=\linewidth]{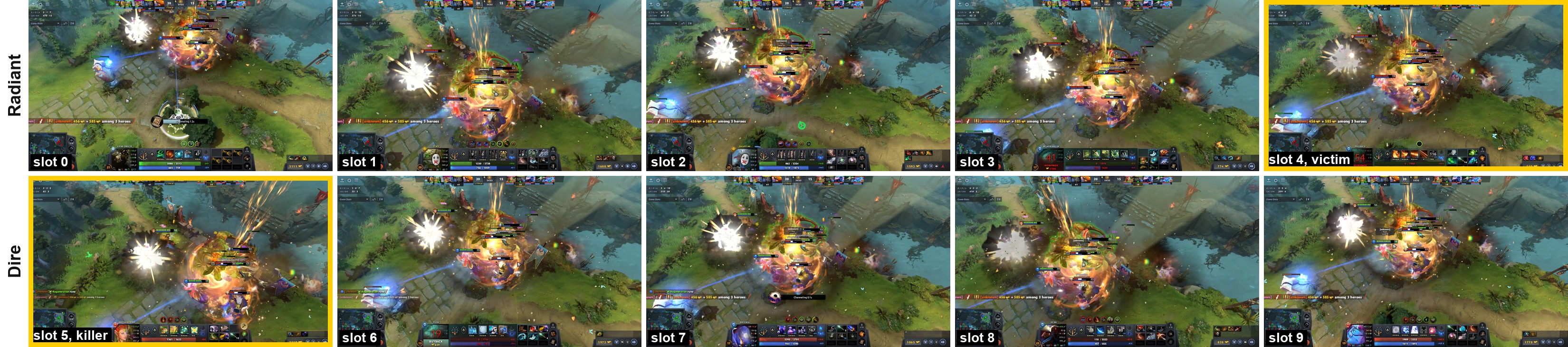}
\caption{Dota~2: all ten Player Views at the replay tick of a hero kill in a team fight (Esports World Cup 2026, match 8893208737), Radiant in the top row and Dire in the bottom row. The victim (slot~4) and the killer (slot~5) are outlined. The released records for this shared tick appear with the enlarged pair in \cref{fig:sample_dota}.}
\label{fig:team_dota}
\end{figure}

%% file: sections/original_visuals/figure_08.tex
\begin{figure}[!htbp]
\centering
\includegraphics[width=\linewidth]{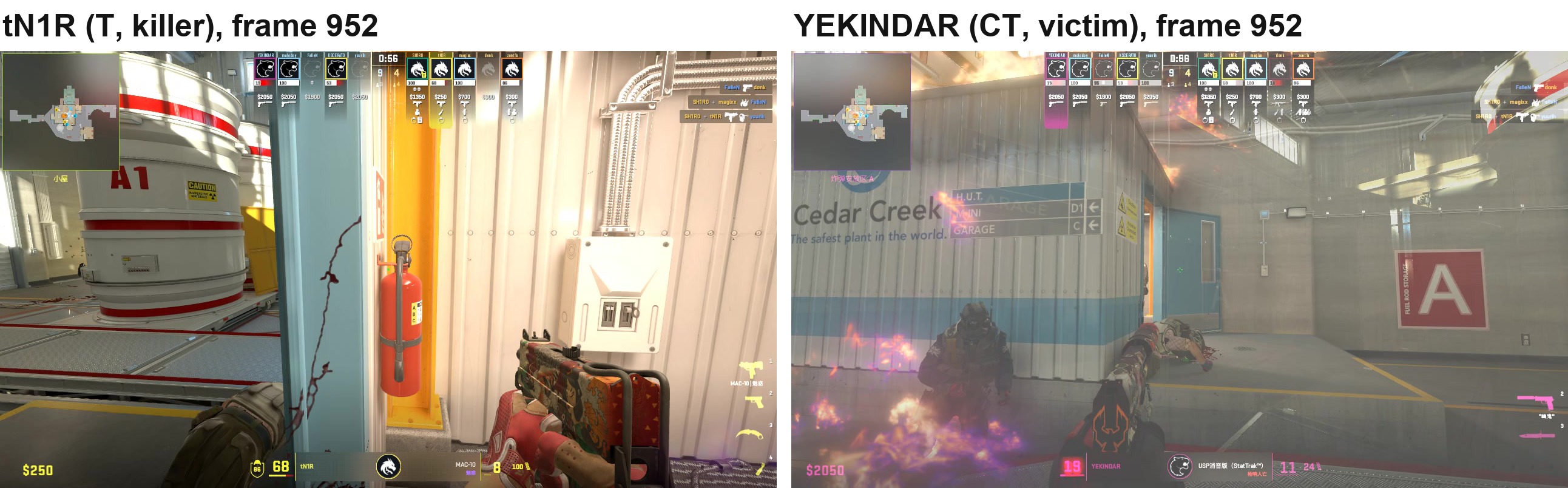}\\[4pt]
\footnotesize
\input{figures/samples/cs2_records.tex}
\caption{Counter-Strike~2: the killer's and the victim's POV in frame 952 of round 14 (IEM Krak\'ow 2026), a quarter second before tN1R kills YEKINDAR with a MAC-10 at tick 132,372. Frame 952 of every POV of the round covers ticks 132,354--132,357; the table lists both players' state at tick 132,356. The two players face each other (yaw $5^\circ$ and $-171^\circ$, 209 units apart along X).}
\label{fig:sample_cs2}
\end{figure}

%% file: figures/samples/cs2_records.tex
\begin{tabular}{@{}llllllll@{}}
\toprule
Player & Side & Tick & Position (X, Y, Z) & Yaw/pitch & Health/armor & Weapon & Buttons held \\
\midrule
tN1R & T & 132356 & (391, -912, -391) & 5/8 & 68/86 & MAC-10 & left \\
YEKINDAR & CT & 132356 & (600, -844, -411) & -171/-4 & 19/0 & USP-S & back, left \\
\bottomrule
\end{tabular}

%% file: sections/new_visuals/team_cs2.tex
\begin{figure}[!htbp]
\centering
\includegraphics[width=\linewidth]{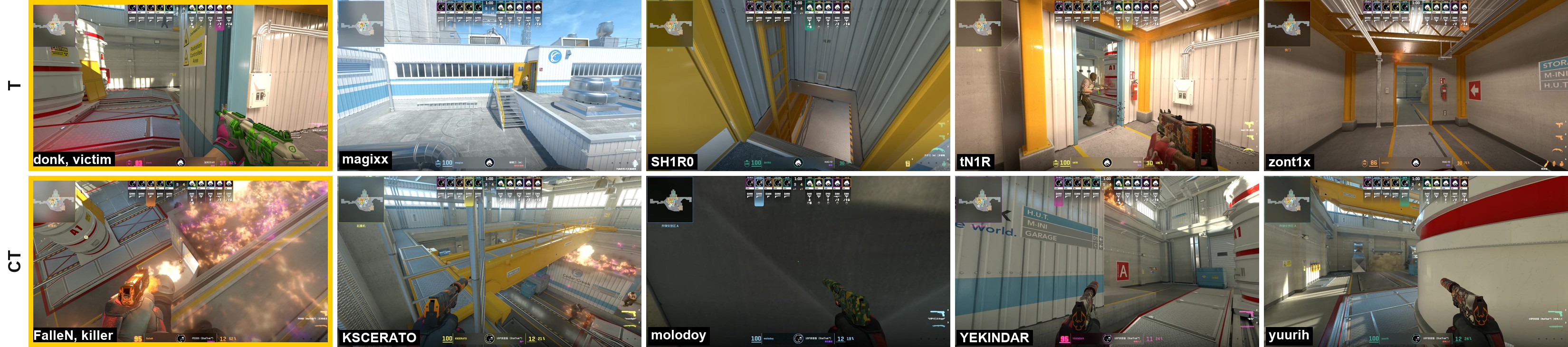}\\[4pt]
\footnotesize
\input{figures/incoming_20261010/samples/cs2_records.tex}
\caption{Counter-Strike~2: all ten POVs in frame 888 of round 14 (IEM Krak\'ow 2026), T in the top row and CT in the bottom row, a quarter second before the first kill of the round: FalleN kills donk with a P2000 at tick 132,117. The killer and the victim are outlined. Frame 888 of every POV of the round covers ticks 132,098--132,101; the table lists both players' state at tick 132,100. FalleN stands 264 units above donk and aims steeply downward (pitch $71^\circ$).}
\label{fig:team_cs2}
\end{figure}

%% file: figures/incoming_20261010/samples/cs2_records.tex
\begin{tabular}{@{}llllllll@{}}
\toprule
Player & Side & Tick & Position (X, Y, Z) & Yaw/pitch & Health/armor & Weapon & Buttons held \\
\midrule
FalleN & CT & 132100 & (322, -826, -128) & -33/71 & 95/0 & P2000 & left, right \\
donk & T & 132100 & (392, -875, -392) & -13/7 & 83/92 & Galil AR & forward, left \\
\bottomrule
\end{tabular}

%% file: sections/original_visuals/figure_09.tex
\begin{figure}[H]
\centering
\includegraphics[width=0.88\linewidth]{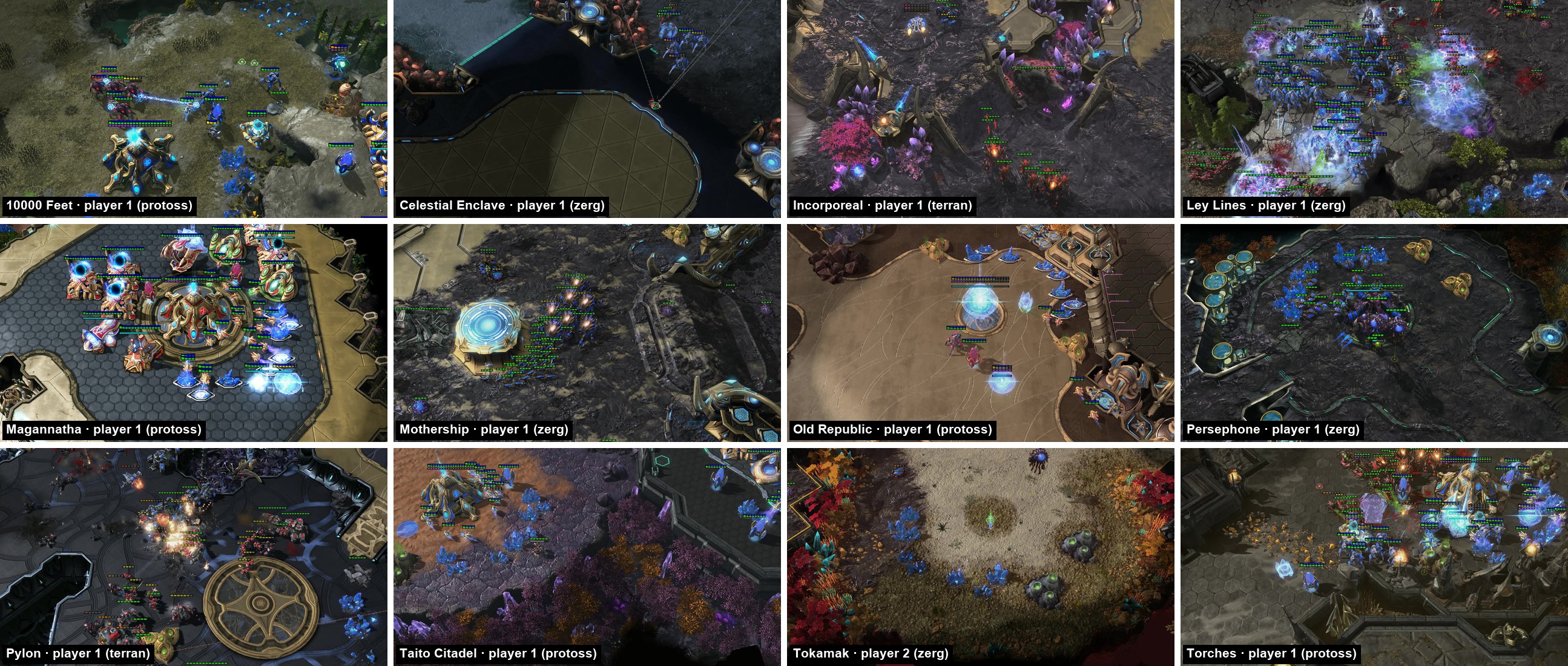}
\caption{\sctwo{} gallery: one game per map, labelled with the map and the player whose camera view is shown.}
\label{fig:gallery_sc2}
\end{figure}

%% file: sections/new_visuals/gallery_sc2_pairs.tex
\begin{figure}[H]
\centering
\includegraphics[width=\linewidth]{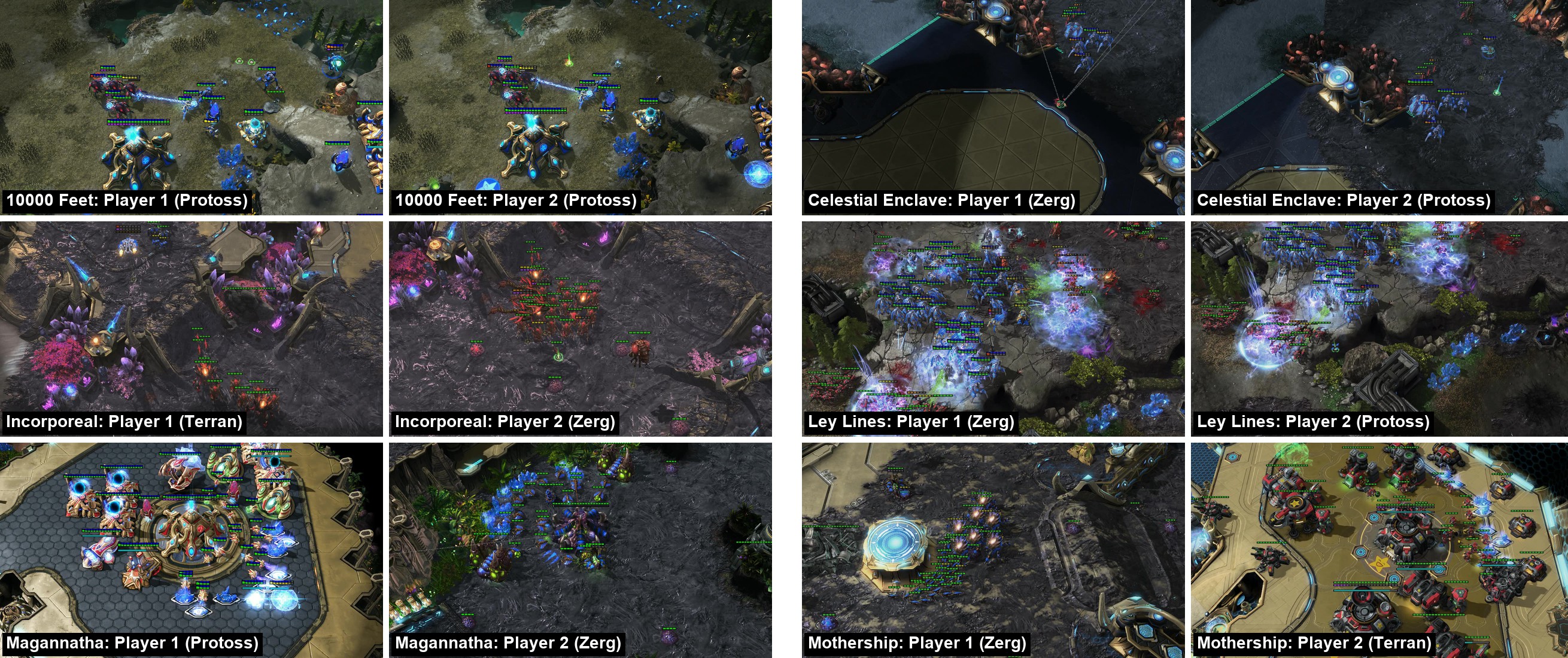}
\caption{\sctwo{} gallery: both players' views at the same game loop, 55\% of the way into one game on each of six maps.}
\label{fig:gallery_sc2_pairs}
\end{figure}

%% file: sections/original_visuals/figure_10.tex
\begin{figure}[H]
\centering
\includegraphics[width=\linewidth]{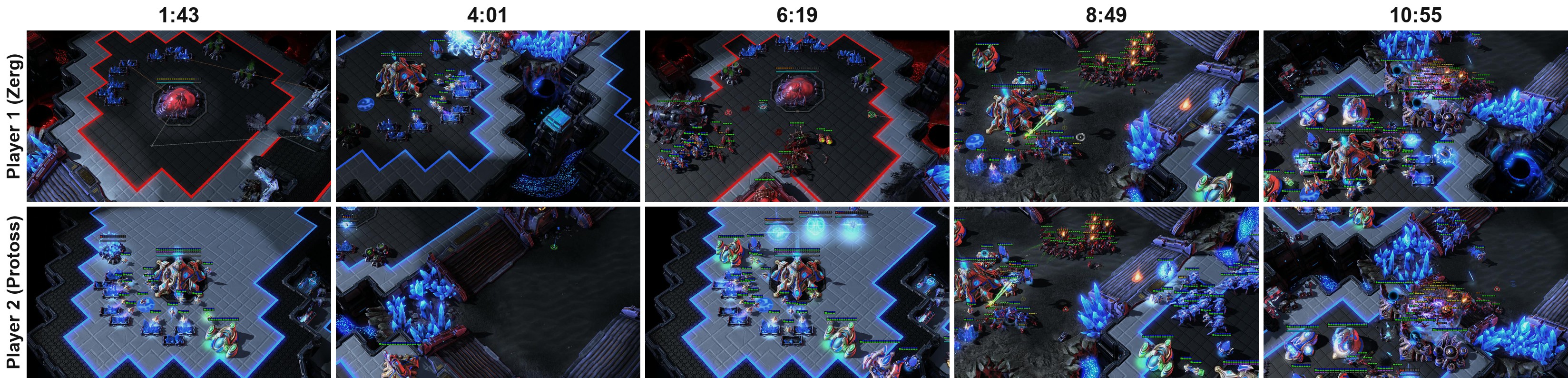}\\[4pt]
\footnotesize
\input{figures/logs/sc2_log.tex}
\caption{\sctwo{} over one game: both players' views at five instants of the Esports World Cup 2025 game of \cref{fig:cover} (Zerg vs.\ Protoss, map Ultralove). The log rows are read from the two trajectories at the same game loops: workers and army supply, unspent minerals and gas, value of enemy units killed so far, and the actions recorded in the preceding minute.}
\label{fig:log_sc2}
\end{figure}

%% file: figures/logs/sc2_log.tex
\begin{tabular}{@{}lrrrrrrrr@{}}
\toprule
 & \multicolumn{4}{c}{Player 1 (Zerg)} & \multicolumn{4}{c}{Player 2 (Protoss)} \\
\cmidrule(lr){2-5}\cmidrule(l){6-9}
Time & Work./army & Min./gas & Killed & Actions/min & Work./army & Min./gas & Killed & Actions/min \\
\midrule
1:43 & 19/0 & 284/24 & 0 & 194 & 20/0 & 25/72 & 0 & 279 \\
4:01 & 36/16 & 93/68 & 0 & 257 & 39/10 & 100/106 & 0 & 338 \\
6:19 & 53/27 & 412/175 & 550 & 333 & 59/31 & 510/477 & 1500 & 330 \\
8:49 & 66/93 & 342/115 & 2700 & 378 & 74/68 & 250/98 & 3275 & 348 \\
10:55 & 65/113 & 447/440 & 6750 & 366 & 78/83 & 860/880 & 8825 & 295 \\
\bottomrule
\end{tabular}

%% file: sections/original_visuals/figure_11.tex
\begin{figure}[H]
\centering
\includegraphics[width=0.88\linewidth]{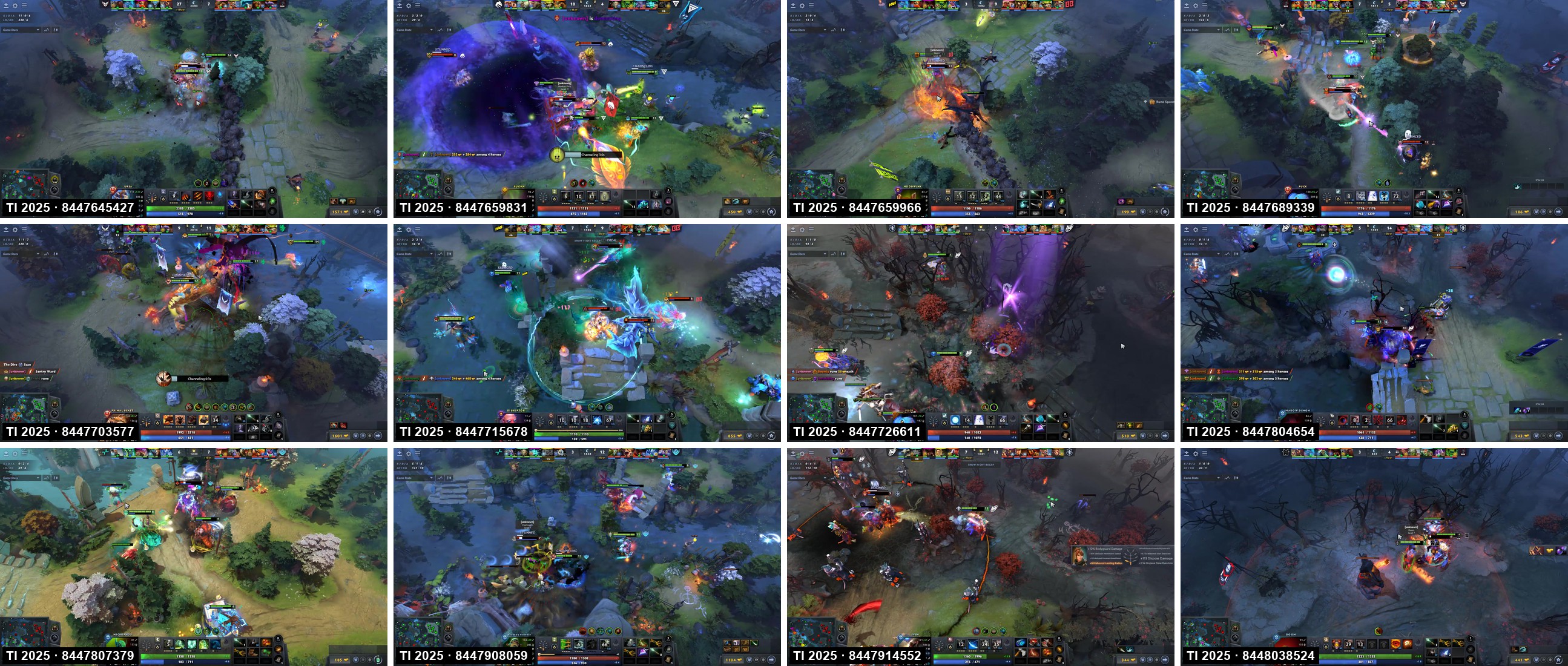}
\caption{Dota~2 gallery: the killer's Player View at hero kills in twelve matches of The International 2025, labelled with the match ID.}
\label{fig:gallery_dota}
\end{figure}

%% file: sections/new_visuals/gallery_dota_team.tex
\begin{figure}[H]
\centering
\includegraphics[width=\linewidth]{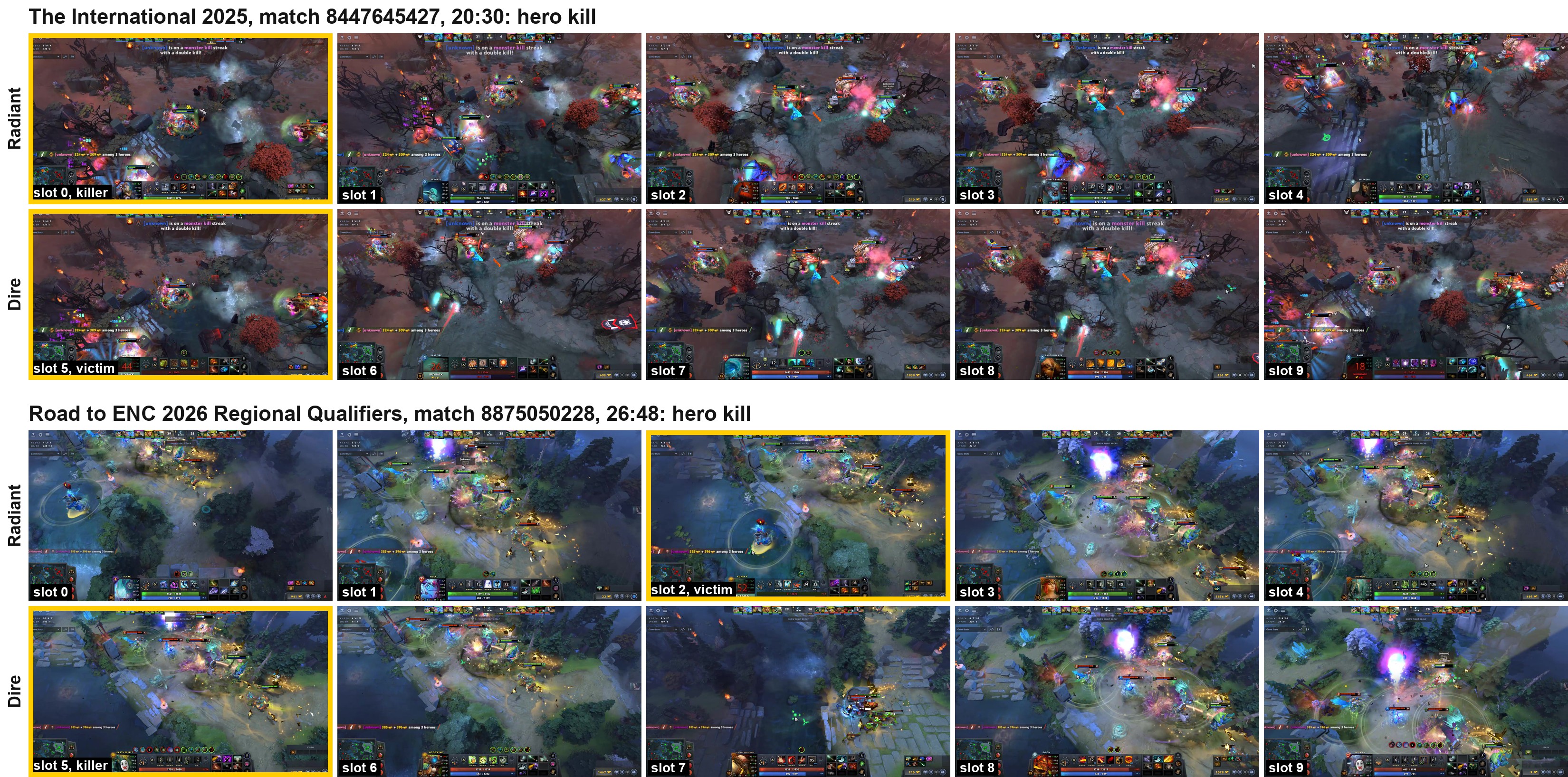}
\caption{Dota~2 gallery: all ten Player Views at a hero kill in a team fight of two further matches (The International 2025 and the Road to ENC 2026 Regional Qualifiers), with the killer and the victim outlined.}
\label{fig:gallery_dota_team}
\end{figure}

%% file: sections/original_visuals/figure_12.tex
\begin{figure}[H]
\centering
\includegraphics[width=\linewidth]{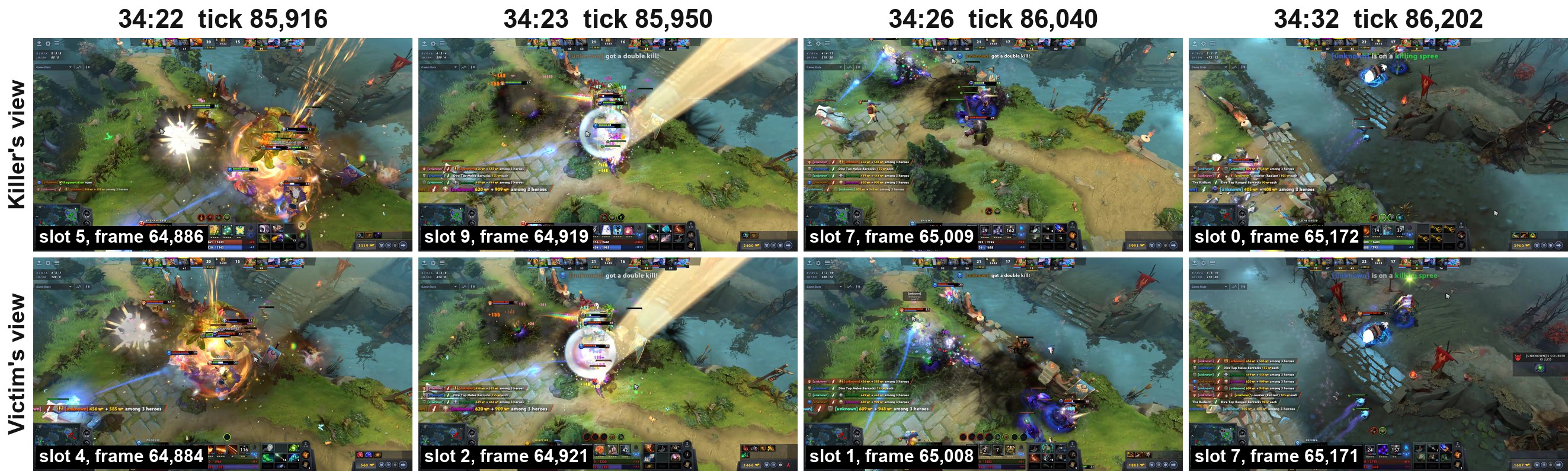}\\[4pt]
\footnotesize
\input{figures/logs/dota_log.tex}
\caption{Dota~2 over one team fight (Esports World Cup 2026, match 8893208737): the killer's and the victim's view at four of the five hero kills, at the replay tick of each kill. The kill log lists all five kills from the chat events, with the gold bounty, the damage dealt by the killer's team in the preceding three seconds (combat log) and the orders the killer issued in the preceding two seconds.}
\label{fig:log_dota}
\end{figure}

%% file: figures/logs/dota_log.tex
\begin{tabular}{@{}rlllrrr@{}}
\toprule
Tick & Clock & Killer & Victim & Bounty & Team damage, 3 s & Killer orders, 2 s \\
\midrule
85916 & 34:22 & slot 5 (Dire) & slot 4 (Radiant) & 456 & 6,591 & 5 \\
85932 & 34:23 & slot 1 (Radiant) & slot 8 (Dire) & 449 & 6,674 & 10 \\
85950 & 34:23 & slot 9 (Dire) & slot 2 (Radiant) & 620 & 6,994 & 16 \\
86040 & 34:26 & slot 7 (Dire) & slot 1 (Radiant) & 609 & 1,296 & 18 \\
86202 & 34:32 & slot 0 (Radiant) & slot 7 (Dire) & 405 & 774 & 13 \\
\bottomrule
\end{tabular}

%% file: sections/new_visuals/log_dota_team.tex
\begin{figure}[H]
\centering
\includegraphics[width=\linewidth]{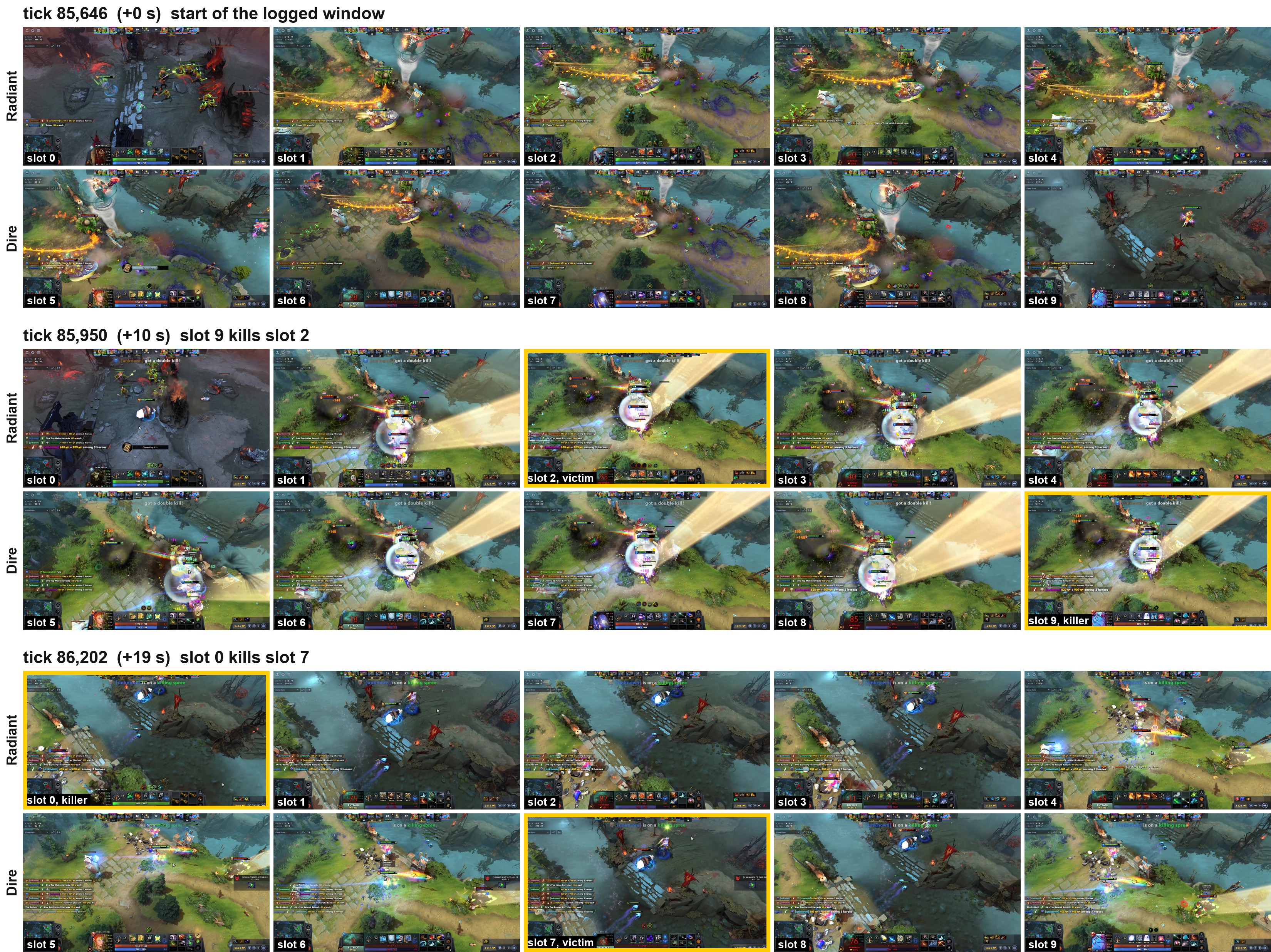}
\caption{Dota~2 over one team fight (Esports World Cup 2026, match 8893208737): all ten Player Views at three replay ticks, the start of the logged window and two of the five hero kills, with the killer and the victim outlined. The kill log in \cref{fig:log_dota} lists all five kills from the chat events, with the gold bounty, the damage dealt by the killer's team in the preceding three seconds (combat log) and the orders the killer issued in the preceding two seconds.}
\label{fig:log_dota_team}
\end{figure}

%% file: sections/original_visuals/figure_13.tex
\begin{figure}[H]
\centering
\includegraphics[width=0.88\linewidth]{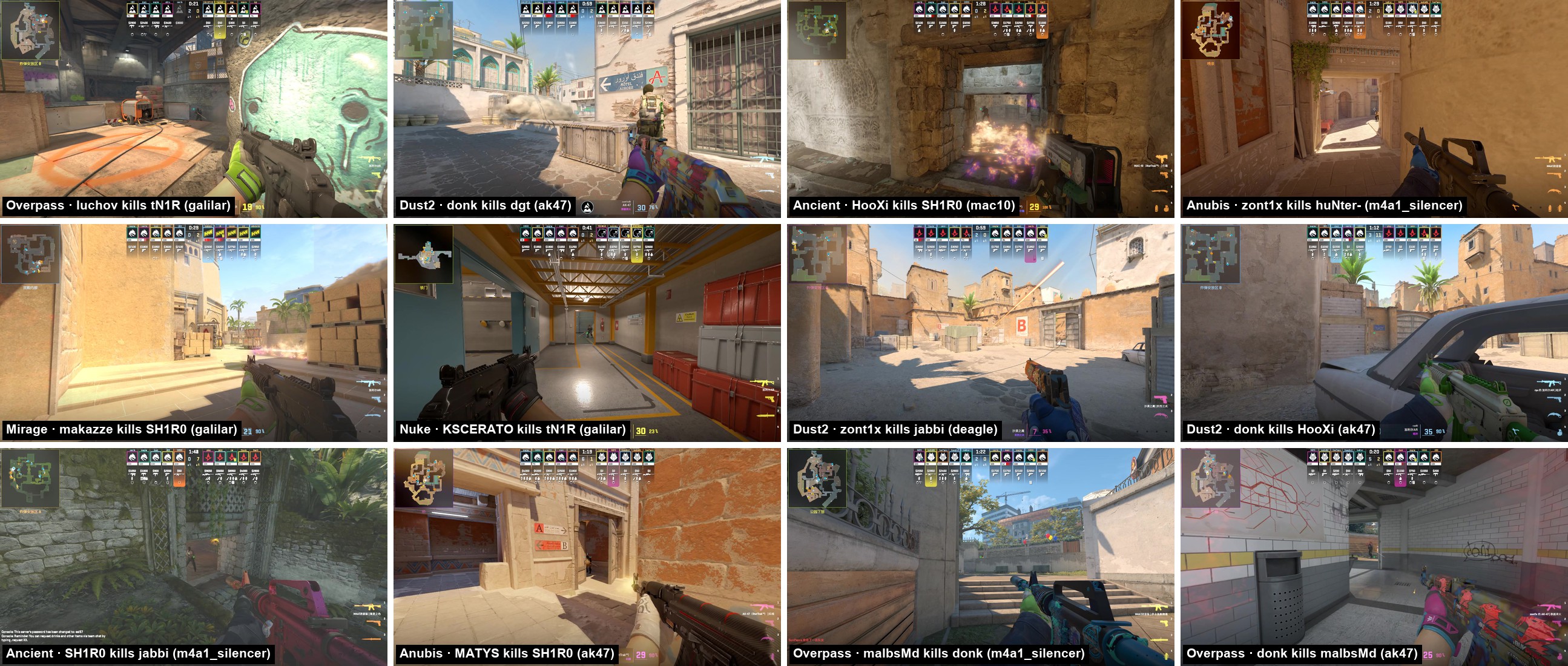}
\caption{Counter-Strike~2 gallery: the killer's POV a third of a second before a kill, on maps of the released series, labelled with the map and the kill.}
\label{fig:gallery_cs2}
\end{figure}

%% file: sections/new_visuals/gallery_cs2_team.tex
\begin{figure}[H]
\centering
\includegraphics[width=\linewidth]{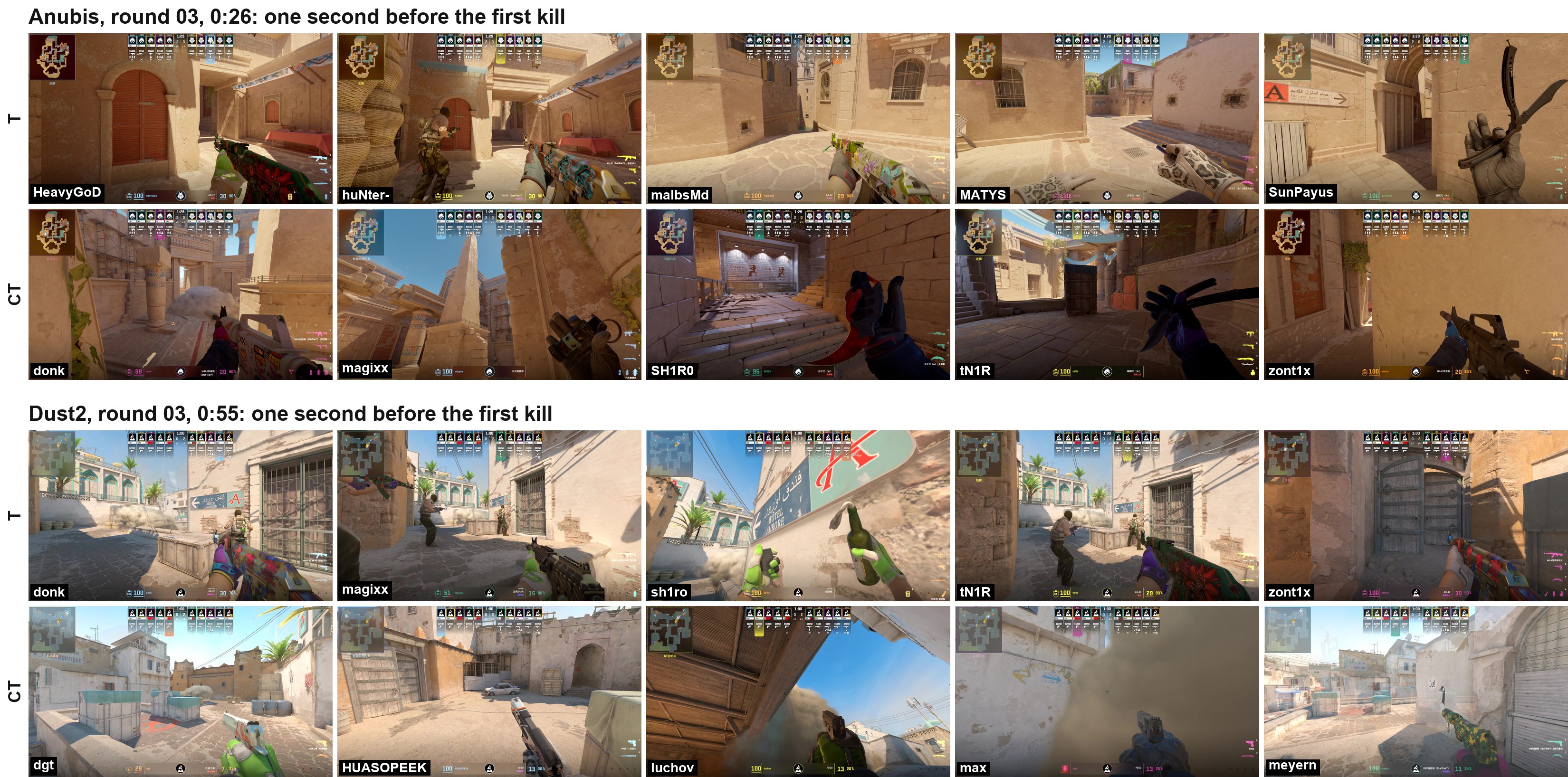}
\caption{Counter-Strike~2 gallery: all ten POVs one second before the first kill of a round on two further maps (IEM Krak\'ow 2026, G2 vs.\ Spirit, Anubis; BLAST Open Rotterdam 2026, 9z vs.\ Spirit, Dust2).}
\label{fig:gallery_cs2_team}
\end{figure}

%% file: sections/original_visuals/figure_14.tex
\begin{figure}[H]
\centering
\includegraphics[width=\linewidth]{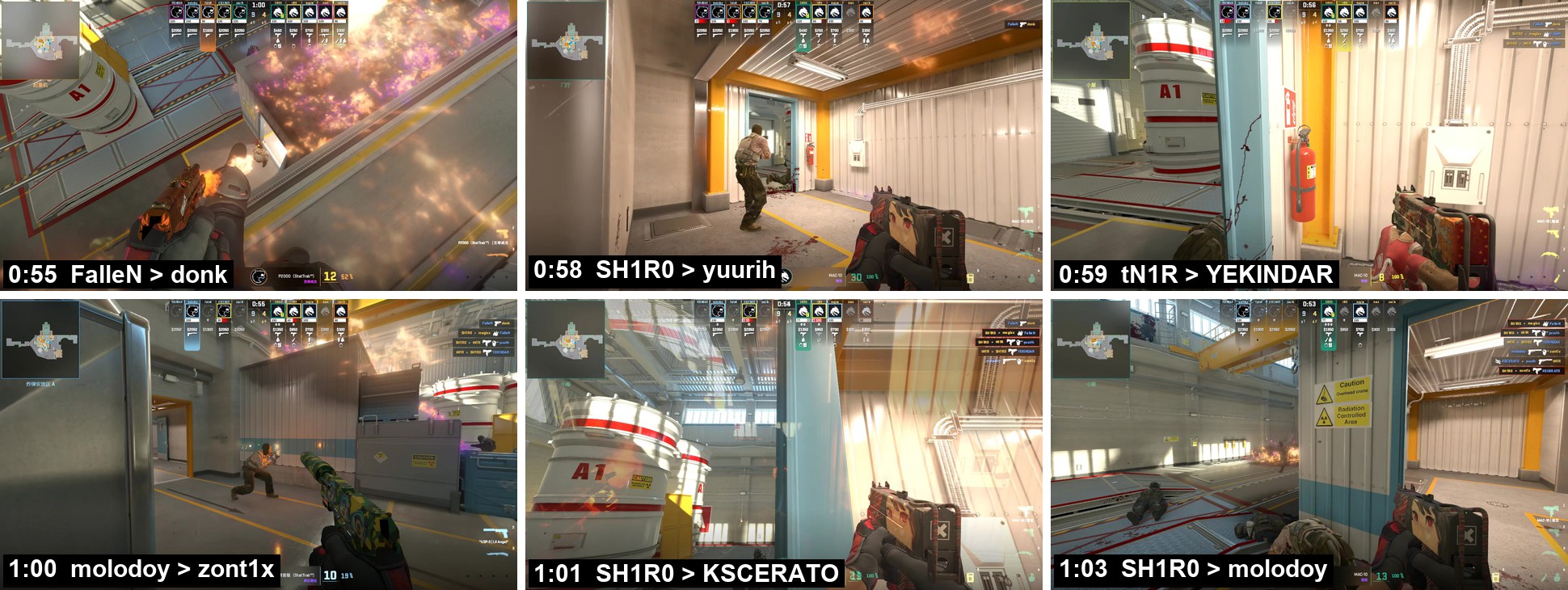}\\[4pt]
\footnotesize
\input{figures/logs/cs2_log.tex}
\caption{Counter-Strike~2 over one round (IEM Krak\'ow 2026, Spirit vs.\ FURIA, Nuke, round 14): the killer's POV a third of a second before six of the seven gun kills. The kill feed is read from the round's event table; the last column is the killer's health in the state table at the kill tick.}
\label{fig:log_cs2}
\end{figure}

%% file: figures/logs/cs2_log.tex
\begin{tabular}{@{}rllllllrr@{}}
\toprule
Tick & Clock & Killer & Victim & Weapon & Hit & Head & Dist. & Killer HP \\
\midrule
132117 & 0:55 & FalleN (CT) & donk & hkp2000 & chest & no & 7.1 & 92 \\
132306 & 0:58 & SH1R0 (T) & yuurih & mac10 & head & yes & 18.3 & 100 \\
132372 & 0:59 & tN1R (T) & YEKINDAR & mac10 & chest & no & 5.6 & 68 \\
132431 & 1:00 & molodoy (CT) & zont1x & usp\_silencer & head & yes & 4.8 & 100 \\
132452 & 1:01 & KSCERATO (CT) & tN1R & usp\_silencer & stomach & no & 16.5 & 6 \\
132473 & 1:01 & SH1R0 (T) & KSCERATO & mac10 & chest & no & 16.4 & 77 \\
132597 & 1:03 & SH1R0 (T) & molodoy & mac10 & head & yes & 9.4 & 63 \\
\bottomrule
\end{tabular}

%% file: sections/new_visuals/log_cs2_team.tex
\begin{figure}[H]
\centering
\includegraphics[width=\linewidth]{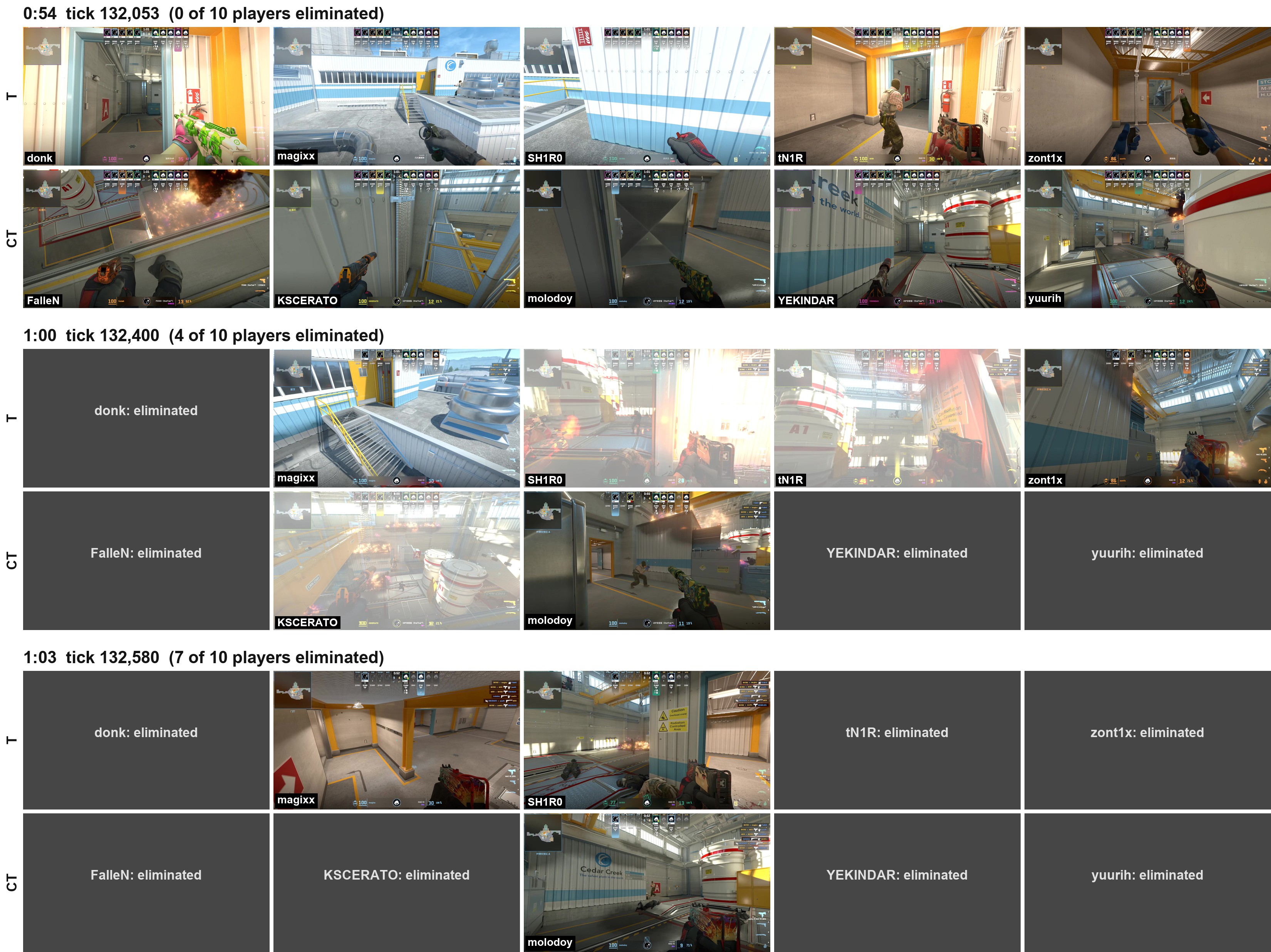}
\caption{Counter-Strike~2 over one round (IEM Krak\'ow 2026, Spirit vs.\ FURIA, Nuke, round 14): all ten POVs at three instants as the round thins out; grey tiles mark eliminated players: a POV is recorded only until two seconds after the player's death (\cref{sec:pipelines}). The kill feed in \cref{fig:log_cs2} is read from the round's event table; the last column is the killer's health in the state table at the kill tick.}
\label{fig:log_cs2_team}
\end{figure}

%% file: sections/original_visuals/figure_15.tex
\begin{figure}[!htbp]
\centering
\recordbox{starcraft2/.../sc2\_5fb15bea5b7c1a55769131f5.p0.trajectory.parquet}{figures/records/sc2_action_compact_rec.tex}\hfill
\recordbox{Player 1 and Player 2 trajectories, observation.units}{figures/records/sc2_state_compact_rec.tex}
\caption{\sctwo{} records. Left: Player~1's trajectory row at the attack command of game loop 12,048 (ability 23, Attack, issued to 28 units towards $(61.47, 113.29)$) and the timing and observation fields of step 12,056. Right: the Archon of \cref{fig:sample_sc2} (box A) nine loops later, in the owner's and in the opponent's observation. The owner's record still carries the commanded target as its queued order; the opponent sees the same tag, position, health and shield but no orders.}
\label{fig:records_sc2}
\end{figure}

%% file: sections/original_visuals/figure_16.tex
\begin{figure}[!htbp]
\centering
\recordbox{dota2/gdrive/matches/8893208737/messages.parquet}{figures/records/dota_action_compact_rec.tex}\hfill
\recordbox{messages.parquet and slot<NN>/alignment.parquet}{figures/records/dota_state_compact_rec.tex}
\caption{Dota~2 records at the hero kill of \cref{fig:sample_dota} (replay tick 85,916). Left: a unit order and the kill message (player~5 kills player~4, 456 gold bounty, game clock 2,062.6~s). Right: the combat-log death of the same kill, and the alignment rows of the victim's and the killer's video, which reach this tick at frames 64,884 and 64,886.}
\label{fig:records_dota}
\end{figure}

%% file: sections/original_visuals/figure_17.tex
\begin{figure}[!htbp]
\centering
\recordbox{counter\_strike2/.../round\_14/76561198872013168/state.parquet}{figures/records/cs2_state_compact_rec.tex}\hfill
\recordbox{counter\_strike2/.../round\_14/events.parquet}{figures/records/cs2_action_compact_rec.tex}
\caption{Counter-Strike~2 records of \cref{fig:sample_cs2}. Left: tN1R's state row at tick 132,356, inside frame 952 of his POV: position, velocity, view angles, health, armor, weapon, money and the buttons held. Right: the kill event 16 ticks later, in which tN1R kills YEKINDAR with the MAC-10.}
\label{fig:records_cs2}
\end{figure}

%% file: sections/new_visuals/records_cs2_team.tex
\begin{figure}[!htbp]
\centering
\recordbox{counter\_strike2/.../round\_14/76561197960690195/state.parquet}{figures/incoming_20261010/records/cs2_state_rec.tex}\hfill
\recordbox{counter\_strike2/.../round\_14/events.parquet}{figures/incoming_20261010/records/cs2_action_rec.tex}
\caption{Counter-Strike~2 records of \cref{fig:team_cs2}. Left: FalleN's state row at tick 132,100, inside frame 888 of FalleN's POV: position, velocity, view angles, health, armor, weapon, money and the buttons held. Right: the kill event 17 ticks later, in which FalleN kills donk with the P2000.}
\label{fig:records_cs2_team}
\end{figure}

%% file: sections/original_visuals/table_03.tex
\begin{table}[!htbp]
\centering
\caption{Scale of the dataset. Timeline hours count each match once; view-hours sum over player views. Structured-data units differ between games.}
\label{tab:stats}
\small
\begin{tabular}{@{}lrrrr@{}}
\toprule
 & \sctwo{} & Dota~2 & \multicolumn{2}{c}{Counter-Strike~2} \\
\cmidrule(l){4-5}
 & & & core & extended$^{c}$ \\
\midrule
Matches & 2,320 indexed games$^{a}$ & 303 & 50 maps & 380 maps \\
Views per match & 2 & 10 & $\le$10 per round & $\le$10 per round \\
Timeline hours & 430.8 & 282.8$^{d}$ & 34.8 & $\approx$265 \\
View-hours & 861.5 & 2,827.9$^{d}$ & 138.4 & $\approx$1,050 \\
Structured data & 69.5M steps & 8.4B state changes & 75.0M rows & $\approx$570M rows \\
Coverage & 18 maps, 3 builds & 381 players, 82 teams & 7,623 clips$^{b}$ & 171 series \\
\bottomrule
\end{tabular}

\vspace{2pt}
\begin{minipage}{0.95\linewidth}\footnotesize
$^{a}$Of the 2,320 indexed games, 2,314 are training-ready and assigned to train/validation/test as 2,088/117/109 games; both views of a game remain in one split.
$^{b}$Per-round clips, one per player.
$^{c}$Hours and rows estimated from the per-map averages of the core set (0.70 timeline hours, 2.77 view-hours and 1.5M rows per map).
$^{d}$Length of the rendered Player View videos, which also cover the pre-game phase before the match clock starts.
\end{minipage}
\end{table}

%% file: sections/original_visuals/figure_18.tex
\begin{figure}[!htbp]
\centering
\includegraphics[width=\linewidth]{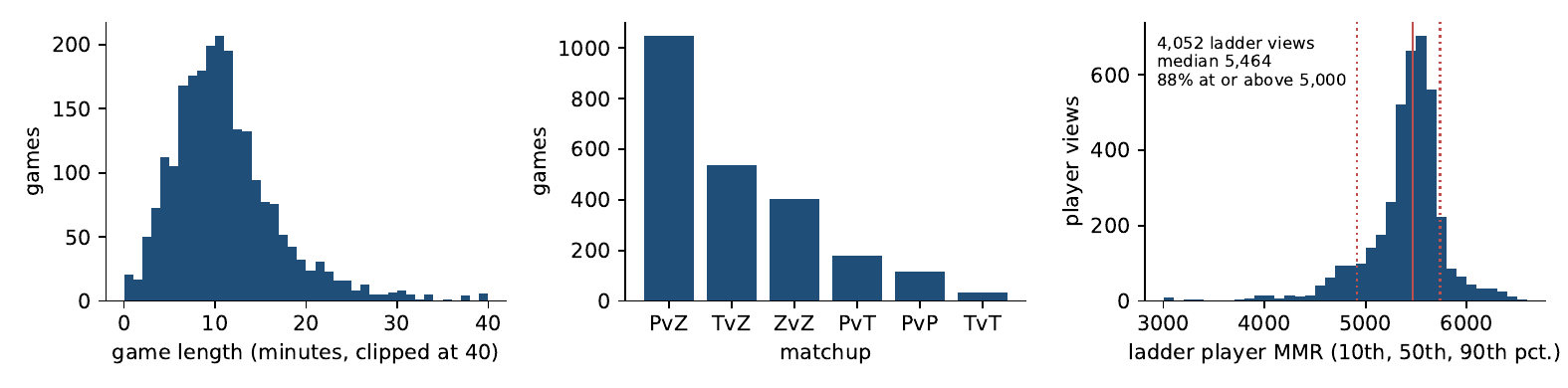}
\caption{\sctwo{} release metadata for 2,320 indexed games and 4,628 completed player views: game length, matchup and player MMR.}
\label{fig:sc2dist}
\end{figure}

%% file: sections/new_visuals/table_events.tex
\begin{table}[!htbp]
\centering
\caption{Sources by event. Units are games (\sctwo{}), matches (Dota~2) or maps (Counter-Strike~2); a series is one best-of match between two teams. Teams and players are distinct teams and player accounts.}
\label{tab:events}
\small
\begin{tabularx}{\linewidth}{@{}Yrrrr@{}}
\toprule
Source & Units & Series & Teams & Players \\
\midrule
\multicolumn{5}{@{}l}{\textit{\sctwo{}} (2,320 games)} \\
\quad Ranked ladder (median MMR 5,464) & 2,076 & & & \\
\quad Rongyi Cup Season~3 (2026) & 90 & & & \\
\quad Kung Fu Cup 2026 & 51 & & & \\
\quad Esports World Cup 2025 & 49 & & & \\
\quad Source not labelled & 54 & & & \\
\midrule
\multicolumn{5}{@{}l}{\textit{Dota~2} (303 matches)} \\
\quad Esports World Cup 2026 & 136 & & 24 & 120 \\
\quad The International 2025 & 96 & & 16 & 80 \\
\quad Road to ENC 2026 Regional Qualifiers & 71 & & 52 & 252 \\
\quad All matches & 303 & & 82 & 381 \\
\midrule
\multicolumn{5}{@{}l}{\textit{Counter-Strike~2, core set} (50 maps)} \\
\quad IEM Krak\'ow 2026 & 12 & 5 & & \\
\quad IEM Rio 2026 & 4 & 2 & & \\
\quad ESL Pro League Season~23 & 3 & 2 & & \\
\quad Seven further events and one unlabelled & 11 & 10 & & \\
\quad All professional maps & 30 & 19 & 17 & 90 \\
\quad Perfect World matchmaking & 20 & & & 175 \\
\midrule
\multicolumn{5}{@{}l}{\textit{Counter-Strike~2, extended set} (380 maps)} \\
\quad IEM Cologne Major 2026 (all stages) & 105 & 50 & & \\
\quad BLAST Bounty 2026 Season~2 (incl.\ Finals) & 54 & 22 & & \\
\quad CS Asia Championships 2026 & 37 & 18 & & \\
\quad IEM Atlanta 2026 & 33 & 14 & & \\
\quad PGL Astana 2026 & 33 & 14 & & \\
\quad XSE Pro League Guangzhou 2026 & 27 & 15 & & \\
\quad Stake Ranked Episodes 2 and 3 & 44 & 18 & & \\
\quad BLAST Rivals 2026 Season~1 & 15 & 7 & & \\
\quad StarLadder StarSeries Fall 2026 EU Closed Qualifier & 14 & 6 & & \\
\quad RES Showdown~4 (South America, Europe) & 8 & 3 & & \\
\quad FRAG St.\ Clair 2026 & 6 & 2 & & \\
\quad Stake Pulse Beat~I & 3 & 1 & & \\
\quad Event not labelled & 1 & 1 & & \\
\quad All maps & 380 & 171 & 61 & 327 \\
\bottomrule
\end{tabularx}
\end{table}

%% file: sections/validation.tex
\section{Data Validation}
\label{sec:quality}

The audits address two prerequisites for using the collection: whether observations can be associated on a shared replay clock, and whether live entity records that refer to the same world state can be compared across views. \Cref{tab:audits} states the scope of each available check and the inference it supports.

\begin{table}[htbp]
\centering
\caption{\textbf{Data audits and alignment checks.} Scope, recorded observations and application to cross-view sampling.}
\label{tab:audits}
\small
\renewcommand{\arraystretch}{1.18}
\begin{tabularx}{\linewidth}{@{}P{0.19\linewidth}P{0.18\linewidth}YY@{}}
\toprule
Check & Scope & Reported observation & Application \\
\midrule
SC2 camera-command timing & 5.6M camera moves & Recorded position matches the command target at the recorded step; shuffled commands give an error of 53.4 map units. & Internal command/observation timing. \\
SC2 co-observed units & 13 games; 5.2M visible-enemy records & Engine-tag joins compare type, position, health and shield. & Comparison of shared entity fields. \\
Dota frame indexing & Six audited matches & Largest replay-time difference at a shared frame index reaches 0.83~s. & Equal frame indices are insufficient for cross-view joins. \\
Dota state reconstruction & Audited change streams & Replaying the deltas reproduces the compared entity-state fields. & Reconstruction of represented entity-state fields. \\
CS2 core alignment & Per-round frame/tick convention & One video frame corresponds to four replay ticks; views share round starts but can end at different times. & Construction of round-aligned cross-player samples. \\
\bottomrule
\end{tabularx}
\end{table}

\paragraph{Clock alignment.}
For \sctwo{}, camera commands provide an internal timing check: across 5.6M moves, the stored camera position matches the command target without a step offset. Shuffling commands within the view increases the reported positional error to 53.4 map units. This check measures the temporal association between stored camera commands and observations.

Dota~2 provides a complementary failure case. Independently rendered views have monotonic frame-to-tick mappings, but the same frame index can refer to different replay times. The largest difference in the audited matches is 0.83~s (\cref{fig:dota}). Joining on stored replay ticks addresses this indexing mismatch. CS2 joins similarly use the round-start tick and the frame interval defined in \cref{sec:alignment}; aligning clips by their ends would introduce offsets caused by different death times.

\paragraph{Live state and remembered observations.}
The SC2 audit compares 13 games at every game loop. It joins 5.2M visible-enemy records to the owner's records by engine tag and compares the common fields identified after \cref{eq:coobserved}. Snapshots are audited separately using their last-seen type and position, because a remembered structure may persist after its live counterpart has disappeared.

The same sample illustrates why multiple views are informative: roughly 10\% of a player's units are visible to the opponent at a time (\cref{fig:visibility}). This statistic describes 26 views from 13 games. The paired observations expose both shared state and substantial private information.

\input{sections/original_visuals/figure_19_20_pair}

\FloatBarrier

%% file: sections/original_visuals/figure_19_20_pair.tex
\begin{figure}[!htbp]
\begin{minipage}[t]{0.485\linewidth}
\centering
\includegraphics[width=\linewidth]{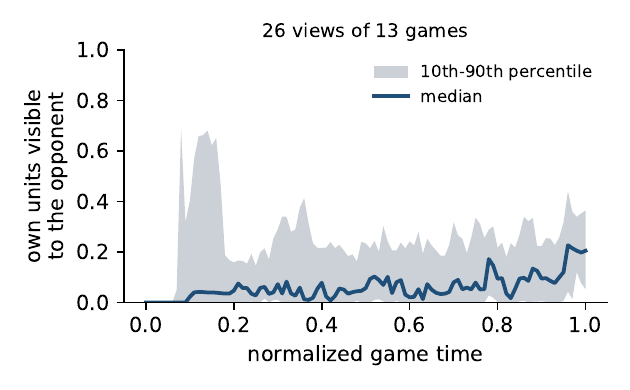}
\caption{Share of a player's units visible to the opponent over a game (\sctwo{}, 26 views of 13 games). Visibility is the engine's fog-of-war state, not camera-window inclusion.}
\label{fig:visibility}
\end{minipage}\hfill
\begin{minipage}[t]{0.485\linewidth}
\centering
\includegraphics[width=\linewidth]{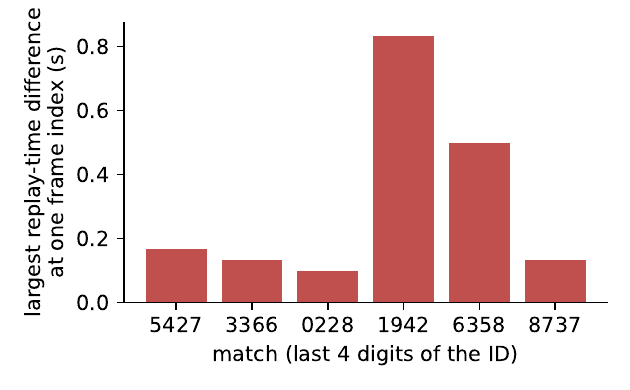}
\caption{Largest replay-time difference between the ten Dota~2 views at one frame index, per audited match. Joining on the replay tick removes this frame-index offset.}
\label{fig:dota}
\end{minipage}
\end{figure}

%% file: sections/discussion.tex
\section{Research Opportunities}
\label{sec:uses}

EWM connects the actions of multiple human participants to their
player-specific observations and the evolution of a shared environment.
The full-team sequences in
\cref{fig:log_dota_team,fig:log_cs2_team} illustrate this connection:
participants experience the same interaction through different views,
with their actions and event outcomes associated on a common replay
clock. This organization provides a data foundation for studying
world models with multiple active participants.

\paragraph{Jointly controlled multiplayer worlds.}
A central direction is to learn shared dynamics conditioned on several
participants' actions while generating observations appropriate to
each player. Such a model could allow multiple users or agents to
cooperate and compete in one evolving world. EWM provides human
interaction trajectories across three complementary settings:
private fog of war, shared team vision and first-person viewpoints.
These settings connect visual prediction to coordination, opponent
adaptation and decisions made under partial information. Cross-view
state agreement and player-specific observation accuracy provide
complementary targets for evaluating the resulting simulations.

\paragraph{Human-data-driven teammates and opponents.}
A complementary direction retains one focal player and learns policies
for the surrounding participants from human trajectories. In a
ten-player setting, one user could interact with nine learned teammates
and opponents, each choosing actions from its own observation history.
The policies determine how participants act, while the world model
predicts the consequences of their joint actions. EWM's attributed
action streams and paired observations provide supervision for studying
both components. This combination could support characters that pursue
objectives, coordinate and initiate encounters, with held-out action
prediction, behavioral diversity and sustained cooperation or
competition providing useful evaluation targets.

\paragraph{Learning under partial observability.}
The same data support predicting a player's future observations and
estimating hidden state from observation history. SC2's live entity
records, remembered snapshots and shared identifiers connect these
tasks to explicit observation semantics. A model can use historical
evidence to infer an unseen opponent's likely state while generating
a view that respects the player's current information. Remembered
snapshots provide historical observations; replay or owner-view
records provide reference state for fields available in those records.
The associated video and action sequences also support long-horizon
video understanding and imitation learning.

\paragraph{Experimental setup.}
Each task should specify its observation modality and action inputs.
Camera RGB, minimap and camera-independent raw observations expose
different information. A focal-player task uses the player's permitted
history and actions, whereas a joint simulator can condition on joint
actions. Dota~2 order tables therefore need to be filtered for
focal-player inputs. Full replay state and other players' private
actions can serve as labels or explicitly privileged training
supervision. Structured evaluation can measure shared-state agreement
and observable-entity recovery; evaluation of video outputs additionally
requires an image-to-entity mapping.

Data should be split before sampling clips, keeping all players,
rounds and clips from a match together. CS2 maps should be grouped by
series when testing generalization across series. The SC2 release
provides the split in \cref{tab:stats}; map-, patch- and player-disjoint
splits address further generalization questions.

\section{Dataset Development and Future Work}
\label{sec:limits}

The current release provides entity-level observation labels for SC2,
replay-rendered views with expert-heavy coverage, and initial audits
on selected matches. CS2 clips end shortly after player death, reducing
late-round view coverage. These characteristics guide the next stages
of dataset development.

\paragraph{Expanding coverage and observation labels.}
The ongoing CS2 collection broadens the available match and participant
coverage. Extending player-specific visibility extraction to Dota~2
would connect its team-view videos to explicit observation labels.
Developing image-to-entity mappings would further link rendered
content to structured records, supporting visual evaluation across
different observation regimes.

\paragraph{Building reproducible benchmarks.}
A next step is to establish fixed splits, input specifications and
baselines for observation prediction, hidden-state estimation and
cross-view consistency. SC2 provides a starting point through its live
observations, remembered snapshots and shared entity identifiers;
comparable protocols can then be developed for Dota~2 and CS2.
Broader data audits should report field-level errors, join failures
and event-to-image timing measurements. Versioned documentation
following datasheet principles~\cite{gebru2021datasheets} can record
the collection, filtering and evaluation procedures, providing a
reproducible foundation for the research directions above.

\section{Responsible Use}
\label{sec:responsible_use}

Non-professional SC2 identities are redacted in the standard release;
identity handling is not uniform across all raw game files.
The project-authored structured data, alignment tables, metadata,
scripts and documentation are provided for non-commercial research
under CC BY-NC 4.0, as described in the release documentation.
Game footage, assets, replays and trademarks remain subject to their
owners' rights and applicable source terms; the project license does
not cover that third-party content. Users should follow the release's
access and reuse conditions. Inquiries may be sent to
\href{mailto:sc2meisah@gmail.com}{sc2meisah@gmail.com}.

\section{Conclusion}

\dataset{} pairs replay-rendered human competitive gameplay with actions and structured records across three observation regimes. These trajectories capture multiple participants whose decisions jointly change one world, with each player observing only part of it. Its central addition is the connection, in SC2, between synchronized player views and entity records that distinguish live shared information from remembered or unavailable observations. Paired encounters, full-team sequences and linked event records show how to construct samples at the level of a participant, an interaction or a shared match. The initial audits illustrate how to inspect these records and why views should be joined on replay time. These checks connect the data organization to the construction of temporally aligned, observation-aware samples. Looking ahead, such data can underpin research on both jointly controlled multiplayer worlds and single-player experiences populated by policies learned from human behavior. The dataset is intended to make those investigations, alongside observation prediction and hidden-state estimation, reproducible and easier to define.